ANOMALY DETECTION IN WIRELESS SENSOR NETWORKS (2.5.6)
Signal Processing and Communications: Project & Thesis - PGEE11110

OLUWASANYA, PELUMI WONUOLA
(s1423789)

# Project Mission Statement

Project Title: Anomaly Detection in Wireless Sensor Networks

Supervisors: Professor Bernard MULGREW

Student's Name: Pelumi OLUWASANYA

PROJECT DEFINITION

A primary motivation for use of wireless sensor networks is change detection over large areas. This might be environmental change such as temperature of the atmosphere or the presence of some characteristic in the received signal that is different from what has been received before. This is anomaly detection, which is, significantly more challenging than conventional detection where we know the signal we wish to detect. Many methods have been proposed for anomaly detection. The one that will be explored in this project is based on estimating the entropy of a signal directly from the data. The entropy is itself estimated through first estimating the probability density function using the k-nearest neighbour (k-NN) technique. This method will be applied to actual measured data.

1) *The pre-dissertation phase will involve a literature survey on the use of anomaly detection within wireless sensor networks as well as preparing a Matlab simulation of suitable test signals*;
2) Simulation and testing of the k-NN technique of [1] on the synthetic test signal evaluated in step 1 to evaluate performance;
3) Verification of the results of [1] on the recorded signals found at: www-personal.umich.edu/~kksreddy/rssdata.html;
4) If time permits, modification of the algorithm to improve performance.

Supervisor: _______________________________

Student: _______________________________

Date: _____________

# Abstract

Wireless sensor networks usually comprises a large number of sensors monitoring changes in variables. These changes in variables represent changes in physical quantities. The changes can occur for various reasons; these reasons are highlighted in this work.

Outliers are unusual measurements. Outliers are important; they are information-bearing occurrences. This work seeks to identify them based on an approach presented in [1].
A critical review of most previous works in this area has been presented in [2], and few more are considered here just to set the stage.

The main work can be described as this; given a set of measurements from sensors that represent a normal situation, [1] proceeds by first estimating the probability density function (pdf) of the set using a data-split approach, then estimate the entropy of the set using the arithmetic mean as an approximation for the expectation. The increase in entropy that occurs when strange data is recorded is used to detect unusual measurements in the test set depending on the desired confidence interval or false alarm rate.
The results presented in [1] have been confirmed for different test signals such as the Gaussian, Beta, in one dimension and beta in two dimensions, and a beta and uniform mixture distribution in two dimensions. Finally, the method was confirmed on real data and the results are presented.

The major drawbacks of [1] were identified, and a method that seeks to mitigate this using the Bhattacharyya distance is presented. This method detects more subtle anomalies, especially the type that would pass as normal in [1].

Finally, recommendations for future research are presented: the subject of interpretability, especially for subtle measurements, being the most elusive as of today.

# Declaration of Originality

*I hereby declare that this report and the work reported herein was composed and originated entirely by myself, at the University of Edinburgh.*

Pelumi Oluwasanya.

## Table of Contents

# List of Abbreviations

| | | |
|---|---|---|
| k-nn | – | k –nearest neighbours |
| pdf | – | probability density function |
| ROC | – | Receiver Operating Characteristic |
| RSSI | – | Received Signal Strength Indicator |
| Q-Q Plot, q-q plot | – | Quartile - Quartile Plot |
| SOM | – | Self-Organizing Maps |
| SVM | – | Support Vector Machines |
| BLE | – | Bluetooth low energy |
| KL Divergence | – | Kullback-Leibler Divergence |
| PCA | – | Principal Component Analysis |

# List of Symbols

| | | |
|---|---|---|
| $f(x)$ | – | Probability density function (pdf) |
| $G(f(x))$ | – | Entropy |
| $\widehat{G}(\hat{f}(x))$ | – | Estimator of the entropy |
| $\hat{f}(x)$ | – | Estimate of the pdf |
| $\widetilde{H}$ | – | Shannon Entropy Estimate |
| $\widehat{H}$ | – | Shannon Entropy Estimator |
| $\mathbb{B}(.)$ | – | Bias |
| $\mathbb{V}(.)$ | – | Variance |
| $\mathbb{E}(.)$ | – | Expectation |
| $Z(.)$ | – | Z-Score |
| $d$ | – | Dimension |
| $o(.)$ | – | Order in probability |
| $f_u$ | – | Uniform distribution |
| $f_b$ | – | Beta pdf |
| $f_m$ | – | Uniform + beta mixture pdf |
| $p$ | – | Mixing Ratio |
| $\mu$ | – | Mean |
| $\sigma^2$ | – | Variance |
| $t_\propto$ | – | Threshold |
| $\alpha, \beta$ | – | Parameters of beta pdf |
| $B(\alpha, \beta)$ | – | Normalising constant for beta pdf |
| $m_i, m_j$ | – | Means of multivariate distributions $i, j$ |
| $\Sigma_i, \Sigma_j$ | – | Covariances of multivariate distributions $i, j$ |
| $B_{ij}$ | – | Bhattacharyya Distance between distributions $i, j$ |

# List of Figures

# Glossary

**Boundary Correction:** This involves reducing the bias due to the k-nearest neighbour (k-nn) region intersecting the boundary of data points close to the boundary by approximating their volume with their nearest neighbour volume whose k-nn region does not intersect the boundary.

**Central Limit Theorem:** Simply put, with sufficient independent realisations of a random variable, its distribution converges to a normal distribution.

**Confidence Intervals:** This is some range of data values where a data point may lie with some specified probability.

**Data-split method:** This involves arbitrarily splitting the data into two, estimating the probability density function from one set, and the entropy from the other set.

**Quartile - Quartile Plot:** This plot compares the distribution of a dataset to the standard normal distribution. If the points lie on a straight line, then asymptotic normality is proved.

**Spatial Anomalies:** These are anomalies that are present in measurements for the same time instant but different sensors or sensor locations.

**Test data:** This is the dataset which the system analyses to detect anomalous measurements using the method that produced favourable results on the training data.

**Threshold:** This is the limit set for normal data point in anomaly detection. Any measurement higher than this is flagged as anomalous.

**Training data:** This is the dataset with which the system learns the normal data values.

# CHAPTER 1
## 1.0 INTRODUCTION
### 1.1 Anomalies

Anomalies are unusual measurements [2] that may be obtained from sensors in a wireless sensor network for various reasons e.g. faulty sensors [3], actual events [4] i.e. a change in some monitored property of the variable, obstructed or even faulty communication system among sensors [5], etc. They represent values, which appear to be different from others obtained from similar ambient conditions in such a way that one is fairly convinced they must be from a different distribution. [6] Anomalies are also called outliers.

### 1.2 Anomaly Detection

Anomaly detection is the process carried out on data points, which ends with specifying whether a data point is normal or abnormal. Anomaly detection methods seek to find the underlying distribution which determines what is normal so as to be able to say exactly (with some level of confidence) when a data point does not follow such. Approaches in the past have been such as to show that an estimator is either biased [1] or unbiased [7] as the number of data points increase. Biases and variances of the estimate are calculated or at least some expression presented [1] [7] [8] [9] [10].

It is worthy of note to mention that anomaly detection is also used to identify replicated nodes, intrusion, e.t.c. Other applications include; Internet traffic anomaly detection where strange data (different from that stored in the server database) sent by a client can be used to identify an anomalous or malicious communication. [11] In another application, e.g. financial, a series of actions initiated by a user can be used to flag a transaction as suspicious if different from what is known as normal. [12]

The goal eventually is to have a system that gives as low as possible false alarm rate as well as very low missed detection. These ensure that the threshold for the hard decision is reached only after some trade-off. The distribution of the estimate is itself expected to play a very significant role in the specification of this hard limit. The methods are highlighted next.

### 1.3 Anomaly Detection methods

Anomaly detection methods are classified into two broad groups. Almost every method belongs to one of these. This division is made based on whether or not there is an underlying assumption in the algorithm. Further sub-grouping is possible, but this is the most common approach. [4] [13] [14]

### 1.3.1 Statistical/Parametric methods

Parametric methods assume a model, and the rest of the work is easier as long as the model is right. This method is based on some knowledge of the data distribution. An important issue is the examination of asymptotic properties of the anomaly detection method. These are usually available for parametric methods in closed forms but not usually in non-parametric methods.

### 1.3.2 Non-parametric methods

Non-parametric models are harder to implement; they seek firstly to find the model (or at least estimate it) before specifying what represents the normal. These are generally more efficient for unknown data than the parametric method.

## 1.4 Approach and metric used for Anomaly detection

This project uses the increase in entropy that occurs when an anomalous measurement is obtained to detect the outlier. This entropy is estimated from the probability density function (pdf) which itself is estimated using the data-split method for k-nearest neighbour (k-nn).

## 1.5 Aim

To verify the anomaly detection method presented in [1] as well as identify its shortfalls and suggest ways of improving it.

## 1.6 Objective
- To estimate the pdf of a dataset using data-split technique
- To estimate the entropy of the dataset.
- To use this metric to detect anomalies.
- To investigate other methods that can be used to improve the result.

### 1.7 Scope of work

This work covers anomaly detection in wireless sensor networks, using the method in [1], which works best on large anomalies, and this work tries to improve on that by presenting a method, which will be sensitive to subtle changes in the pdf of the dataset.

### 1.8 Outline of the report

This report is arranged as follows, chapter two opens with a brief peek into wireless sensor networks and concludes with a review of some related works in anomaly detection for wireless sensor networks. Chapter three presents a description of the method of [1], with a test of the method on data from known distributions and in different dimensions to show how it performs or whether some application of the method may be found for such distributions while recognising the drawbacks. Chapter three concludes with a verification of the method on real data. Chapter four presents a new approach based developed based on the method of [1] but dealing with some of its most important issues. This method is shown to produce better results than [1]. Chapter five summarizes the main points again and gives an insight into what may be considered for future research.

# CHAPTER 2
## 2.0 BACKGROUND

Since the release of the book by D. Hawkins titled 'Identification of Outliers' [15], anomaly detection has been approached from different directions. Some of the works that have been done have been discussed in [2] and would not be repeated here. The reader may find it useful to read it to get the writer's critical review of those works as only other works are discussed here. It should be stated that the books by Aggarwal [6] and Hawkins are not limited to wireless sensor networks as this work is and this difference should be noted. This chapter presents a background on the subjects of wireless sensor networks, anomalies and some related works in anomaly detection.

### 2.1 Wireless Sensor Networks

Wireless sensor networks are usually a set of sensors deployed to monitor changes in their environment. [16] Depending on the size of the area where these variables are to be monitored, the number of sensors deployed may vary. From just a few sensors to a very large number of sensors all measuring and reporting some variable(s).

These sensors are usually powered by batteries, and may require some maintenance. The sensors communicate through a wireless link, which may be achieved by using the IEEE 802.13 standards. These are usually better used in instances where power conservation is not the most important requirement as there are now more energy conserving links such as the Bluetooth and the more recently discovered Bluetooth low energy (BLE). Different protocols exist and many more are being developed to tackle fierce challenges, and research is currently going on worldwide to continually beat the energy requirements of the sensor network down as much as possible. Sensors are also being designed to achieve this [17]. It may be useful to mention that one of the most likely causes of this is the application of wireless sensors in healthcare. Humans now wear sensors and thus newer approaches have to be found first to make it safe as well as make it more effective. Of course size becomes an issue when these are considered, and thus sensors can only continue to get smaller and smaller, yet they must be more effective. Drug delivery and artificial organs, etc., are some other areas that sensors are being developed. Also, it has become very important that sensors in remote areas do not require maintenance, such as replacement of batteries, too often.

In all these, sensors, no matter their number, measure variables, communicate it to some location, and these are interpreted as information. This information can be lifesaving, at times e.g. in the prediction of a natural disaster, or in the prompt alerting of healthcare practitioners to an ailing patient. Data provided by sensors are very important. If there is anything that should be jealously guarded, it probably is the integrity of measurements returned by the sensors. There should be no doubts concerning their authenticity because life-defining decisions may be made based on these measurements and it may be unforgivable to do that on faulty sensor data. This is because it is our sincere wish to take any changes in measurement as an actual event. But sadly this wish seems very hard to achieve for several reasons. Most of these have already been identified in [2]. Sensors may return strange measurements when their batteries are low, some may intermittently turn on/off in this condition and this may even result in missing measurements, these are usually grouped under faulty sensors. Sabotage, apart from node replication, may take the form of intentionally incrementing measurements, or outright destruction of the sensor nodes or their motes.

Having already mentioned that anomalies occur and now that they are almost inevitable, it becomes imperative to subject measurements to some test to identify the strange ones as well as to know why they are strange. Only on successful verification would it be safe to make decisions based on such measured data. Anomaly detection has therefore, come to stay.

The following types of anomalies are identified based on some of these causes.

## 2.2 Types of Anomalies in Wireless Sensor Networks

### 2.2.1 Extreme values

Extreme value anomalies are the usual anomalies that occur when a sensor turns on or off arbitrarily while taking a measurement or a sudden surge of current through the sensor. A common value for this is zero or an excessively large value.

### 2.2.2 Missing Data

Missing data may result in the sensor defaulting to zero, and are similar to extreme values except that they do not take up large values, and may be even marked as unavailable altogether.

### 2.2.3 Constant/Variable Increment/Repetition

Anomalous measurements due to sabotage may take the form of deliberately inflating figures either by a constant amount or by a variable amount or even intentionally repeating data through node replication or node cloning. These are usually due to sabotage or malicious intentions.

### 2.2.4 Subtle anomalies

Subtle anomalies usually represent the form when a large change happens very slowly and may escape notice or when just a little change happens and then there is a return to the norm. This is quite harder to detect and may sometimes represent an actual event such as in the case where there is a gradual reduction in functionality or overall health in medical applications.

## 2.3 Causes of Anomalies in Wireless Sensor Networks

### 2.3.1 Faulty sensors or motes

Broken sensor, dead batteries, non-deliberate obstruction of wireless sensor communication, faulty motes, etc., all qualify as anomalies caused by faulty sensors or equipment. This is especially true if the anomalies disappear once the faults are fixed, if not, sabotage will be the more likely cause.

### 2.3.2 Sabotage

For this, an enemy that seeks to water down, or aggravate measurements with an aim to mislead decision makers causes the anomalies deliberately. This usually may take several forms. This has led to the research in the area of security in wireless sensor networks, with different already developed. [18] [19]

### 2.3.3 Errors

Errors may be due to changes in sensor intrinsic characteristics. An example is the changes in measurements by a sensor at different temperatures. Errors may also occur when the sensors communicate non-steady state measurements. [20]

### 2.3.4 Events

Events are the actual anomalies that the wireless sensor network designer wishes to have to deal with. They represent the actual information. For whatever application, events are the changes that need to be monitored and should affect decision-making processes that take place based on measurements from wireless sensor networks.

## 2.4 Some Related Works

Here are some of the works done in this area that were not already mentioned in [2] which may serve to shed more light on the ideas from the last section.

[21] presents an anomaly detection method in a wide area wireless sensor network where sensor nodes form clusters around specific smaller areas in the network and sensor measurements are allowed to vary from cluster to cluster. The authors use an algorithm which comprises two stages: the data correlation stage and a sketch stage. In the first stage, sensor physical positioning is used to divide the measured values into clusters and the second stage then uses the kullback-leibler divergence to compare measurements from each sensor in a cluster with others by the cluster head. There are results in the paper which suggests a robust performance of the technique and it is quite intuitive except that the authors say it has not been tested on actual sensors and also another look at how the method works suggest that the performance may not be as desired in a situation when sabotage was intended, and the entire cluster was attacked. In fact only a majority need to be attacked before the system sees the anomalous measurements as the normal and the normal as anomalous. Furthermore, while the KL divergence is a good metric, the effect of subtle differences in an entire dataset may be lost for a very large number of measurements. This work will try to solve this problem. It is also not clear how this method performs for higher dimensional data and no mention was made of the complexity costs of the method.

[22] is an overview of different approaches to anomaly detection that the authors consider as state of the art and would be a good read for someone who wants to have a single material that covers as much of the research area as possible. Statistical techniques and non-parametric techniques are presented as the major divisions, as do some other works as shown in section 1.3.

In [23], an application of wireless sensor networks to healthcare is presented, and an anomaly detection is developed not only to detect a strange measurement but also to specify if the cause was a faulty sensor or an actual failing health. This is then used to know who to alert, e.g. doctors, etc. This technique is also in two stages. In the first stage, a decision tree is used to differentiate normal from abnormal data points while a regression is used to determine whether the abnormal measurement was due to a fault or an actual change in health conditions of the patient. This is a good technique as it is one of the very few works that have attempted to discuss the interpretability of anomaly detection results, and that with low false alarm rate. The importance of this may be stressed by saying that it is better and time-saving, and maybe even life-saving to have a technician, rather than a doctor respond to faulty sensor issues; of course except if the sensor is implanted. Simulation results presented in this work look promising.

The method in [24] relies on the interpretation if sensor data as signals that can be projected into subspaces pertaining to normal and abnormal conditions using some orthogonal matrices. It appears to have been developed to surmount some of the challenges of a principal component analysis (PCA) identified in the work and is shown to outperform it in terms of false positives, and in that it can be used for distributed anomaly detection. The reader is however reminded that projecting data to such normal and abnormal subspaces assumes that such subspaces always exist. This may not always be the case.

[25] was developed to detect anomalies that are caused by sabotage. In this case, the enemy physically moves the sensor(s) from its location. The motivation for this by the enemy is mostly based on avoiding discovery, and deception. This movement results in sensors returning measurements that do not represent their location as expected. The fundamental assumptions in this work are that the enemy may not move most sensors in the network at the same time, and that the initial position of the sensors are known to each sensor, e.g. by broadcasting, and that the sensors are static. The key to this work lies in the fact that received signal strength indicator (RSSI) of a sensor signal varies with position and the sensors are static sensors. This is used with a log-normal shadowing model to evaluate changes in sensor position from its RSSI. This method is quite robust for its application because it uses a cooperative approach to concluding on an anomalous node. Sensor measurements are compared for sensors from time to time and when a sensor is suspected to be anomalous, its neighbours have to agree to same by returning a change in the estimate of its location, but more interesting is the fact that the anomalous node will also return a change in location for all its neighbours. The method is good provided its assumptions hold true else, it no longer holds much water.

The method in [26] was developed as an improvement to anomaly detection using self organizing maps (SOM) with support vector machines (SVM). An SVM based approach similar to this has already been discussed in [2]. The authors identify the complexity of the calculations as one of the drawbacks, among others. Their solution uses the Stockwell transform or S-Transform in combination with SVM. The S-Transforms allows for the extraction of four significant features that are used for the detection. This is the first stage. It separates the data into its time and frequency components. The SVM stage is used to classify the training data into two groups; normal or anomalous. This method reduces training time also. Training data was generated from humidity and temperature sensors while anomalies were deliberately introduced by adding steam from a kettle. From the work, it is however not exactly clear how to select the features or why they were selected, and the authors reported that one of the features did not give satisfactory results, while the others do quite well.

### 2.5 Conclusion

This chapter provides a brief background on wireless sensor networks and introduces the problem of anomalies in these networks, identifies its causes as well as its types. It ends with a critical review of some related works that were not already reviewed in [2] in preparation for the methods of chapter 3 and 4.

**CHAPTER 3**

**3.0 METHOD**

### 3.1 Entropy

Entropy is the average information present in a dataset. [27] When datasets of the same values are received one after the other, there is no information in the second dataset that cannot be obtained from the first. But when anomalies are present, datasets differ and there is some information. This causes an increase in entropy. This increase in entropy with every anomalous measurement recorded is used to detect the anomalies in this work. Put in another way, the amount of bits required to represent the dataset increases when outliers occur. In its simplest terms, this can be interpreted as meaning that it is easier to say ten eggs than to say nine eggs and one fish [6]. Here is what we now know for certain; if there is a way we can measure or keep track of changes in entropy with each dataset obtained we may be able to say, within limits of some error, when an anomalous data point is returned.

### 3.1.1 Shannon Entropy

Shannon Entropy is defined as:

$$\boldsymbol{G}(f(x)) = \int_{-\infty}^{+\infty} -f(x)\log(f(x)) \qquad (1)$$

Where: $f(x)$ is the probability density function of the dataset.

There are two main issues with applying (1) to the problem;

1. We do not have the $f(x)$ since we know nothing about the dataset except that we have it, and that it is just a set of numbers. Thus we need to evaluate it or at least approximate it in a reasonably accurate manner.
2. Equation (1) presents a continuous integration, however all we have is a set of discrete numbers which do not even run from beginning of time to the end of time $(-\infty, +\infty)$, and thus $G(f(x))$ will only be approximated as;

$$\widehat{\boldsymbol{G}}(\hat{f}(x)) = \frac{1}{N}\sum_{n=1}^{N} -\log(\hat{f}(x)_n) \qquad (2)\ [1]$$

Using dataset of size $N$.

This comes with its own problem: more and more data will be needed to get more accurate result and thus the degree of accuracy to be achieved would have to be decided based on some trade-offs.

The first issue can be solved in a number of ways but in this project, the *k*-nearest neighbour approach is used. It is however not without its own problem, as will be seen, boundary problem for bounded dataset is one of such and an attempt to solve this problem is also presented.

## 3.2 k-Nearest Neighbour method of PDF Estimation
### 3.2.1 Definition

This method uses the distance between a data point and its *k*th nearest neighbour to calculate the volume of the region used to approximate the pdf of the dataset at that point from the equation:

$$\hat{f}(x) = \frac{k-1}{NVol} \qquad (3)\,[1]$$

Where: $Vol$ is the volume of the *k*-nearest neighbour region for each data point, $k$ is the number of nearest neighbours considered, and as before, $N$ is the size of the dataset. The reader should note that the distance measure used is the Euclidean distance and also equidistant in all directions. This is a circle in two dimensions, sphere in three dimension, etc.

But another problem quickly comes to the fore; Equations (2) and (3) are coupled! [9] This may have some impact on the result. To avoid this, a data-split method is invoked.

### 3.2.2 Data-split method for k-Nearest Neighbour

This entails splitting the given dataset arbitrarily into two. These can be arbitrarily labelled $N$ and $M$. From the $M$ data points the pdf of the entire set is estimated, the $N$ data points are then used to estimate its entropy. Equation (3) is then written as:

$$\hat{f}(x) = \frac{k-1}{MVol} \qquad (4)$$

This has been shown in [2] to provide better result than using the entire set.
Also, as will be shown, the more the points used in estimating the pdf for the $M$ data points, the better the estimate. The reader may observe that in (2), all the terms retain their meanings except that now $N$ represents a part/section of the entire dataset and not the whole set as may be expected. With this knowledge, (2) can be referred to without ambiguity or loss of meaning.

### 3.2.3 Boundary Correction for data points close to the boundary

In the two dimensional data shown in the figure below from a beta distribution with parameters (4, 4), data points near the boundary on each axis will have the distance to their $k$-nearest neighbours greater than their distance to the axis, thus the volume of the $k$-nn region intersects the axis. This introduces errors in the estimate of the actual volume of the region, this error is undesirably carried on to the pdf estimate.

An approach presented in [9] to circumvent this problem is to assign to each point with this problem, the pdf of its nearest neighbour whose volume does not intersect the axis. This is of course based on the assumption that the pdf is uniform and equal at both points. The error incurred by doing this can be shown to be lesser than that incurred if no boundary correction is applied. Figure 3.1 shows such data points.

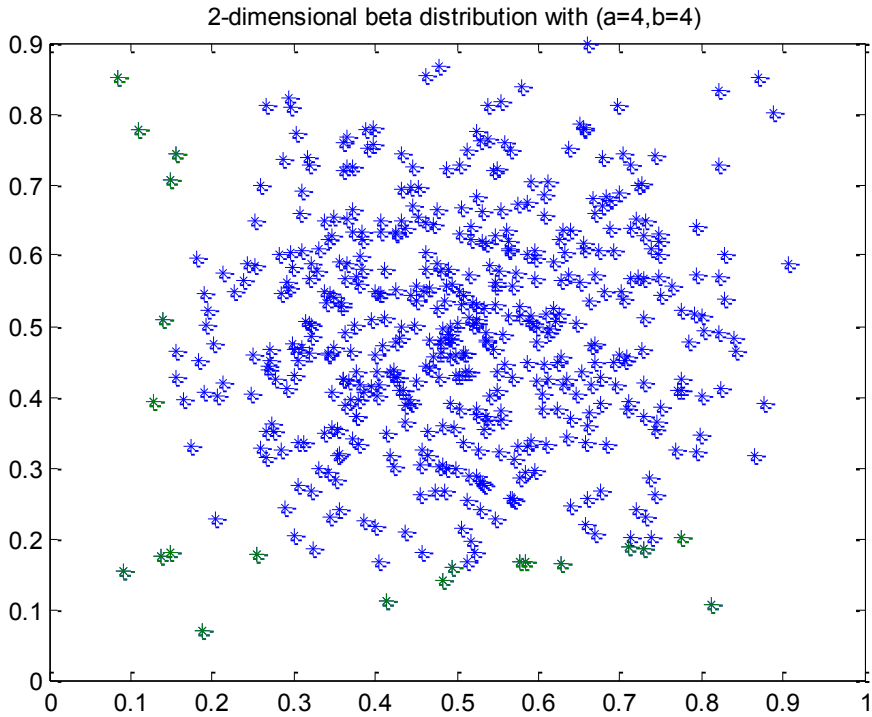

*Figure 3.1 Data points near the boundary are shown in green, their k-nn region intersects the boundary.*

### 3.3 Functionals, Estimators and Estimate

In this project, the entropy functional to be estimated is of the form:

$$G(f) = \int_{-\infty}^{+\infty} g(f(x))f(x)d\mu(x) = \mathbb{E}[g(f(x))] \qquad (5)\,[1]$$

Where $\mu$ represents the Lebesgue measure, and $\mathbb{E}$ is the expectation. According to [1] on which this work is based, the choice of functional for the entropy is $g(x) = \log(x)$. The pdf estimator is the $\hat{f}(x)$ which is evaluated at every point in the dataset of interest to approximate the pdf of the set. The entropy estimate obtained from the entropy estimator of (2) can then be written as:

$$\widetilde{H} = \widehat{H} + [\log(k-1) - \psi(k-1)] \qquad (6)[1]$$

### 3.4 Plug-in Estimator

The term 'plug-in estimator' is used when referring to the Shannon entropy estimator derived by applying (2) and (4) to a dataset because the function $\hat{f}(x)$ obtained using (4) in the first part of the dataset is then 'plugged-in' to (2) to obtain the Shannon entropy estimate using the second part of the dataset. So it seemingly appears to be two different datasets for different functions one needing the output of one to run appropriately. But it is still the same dataset and the reason for this as already heighted is to prevent coupling.

It should also be noted from (2) that the pdf estimate must be bounded away from zero and infinity for all values of $x$, i.e. ($0 < \hat{f}(x) < \infty$). With this in mind one would expect to have the results from [1] hold for only such pdf.

### 3.5 Main Findings
#### 3.5.1 Bias and Variance of the Estimator

The bias of the plug-in estimator has been shown in [1] to be;

$$\mathbb{B}\left(\widehat{G}(f)\right) = c_1 \left(\frac{k}{M}\right)^{\frac{1}{d}} + c_2 \left(\frac{1}{k}\right) + o\left(\frac{1}{k} + \left(\frac{k}{M}\right)^{\frac{1}{d}}\right). \qquad (7)$$

Furthermore, the variance of the plug-in estimator was found to be:

$$\mathbb{V}\left(\widehat{G}(f)\right) = c_4 \left(\frac{1}{N}\right) + c_5 \left(\frac{1}{M}\right) + o\left(\frac{1}{M} + \frac{1}{N}\right). \qquad (8)$$

Proofs of equations (7) and (8) are shown in the appendices of [9] and would not be repeated here.

### 3.5.2 Asymptotic Normality using Central Limit Theorem

It is shown that when the plug-in estimator of the entropy is properly normalized, it converges somewhat weakly, to a normal distribution. This result is however key in application to anomaly detection especially because it makes it much easier to set thresholds using confidence intervals. Asymptotic behaviours for the following conditions will be considered: (i) $\frac{k}{M} \to 0$, (ii) $k \to \infty$, and (iii) $N \to \infty$. More conveniently written as $\Delta \to 0$, and the result presented as:

$$\lim_{\Delta \to 0} Pr\left( \frac{\widehat{G}(f) - \mathbb{E}[\widehat{G}(f)]}{\sqrt{\mathbb{V}[\widehat{G}(f)]}} \leq \alpha \right) = \Pr(Z \leq \alpha) \qquad (9)[1]$$

Where $Z$ is a standard normal random variable. This will be shown via the q-q plot.

### 3.5.3 Optimality of $k$

This best value of $k$ is defined based on how the pdf estimate improves with increasing $k$ until when there seems to be little improvement in the result despite increasing $k$. So this will generally be subjective and some compromise may have to be reached on how much is enough considering constraints like complexity and calculation costs. It is application dependent. In this report, the rule of thumb, which says that the square root of the dataset size is optimal, is found to be true as shown in some of the results where the pdf estimate obtained using this value of $k$ doesn't improve too much as the value of $k$ becomes larger. Even though the pdf then smoothens out.

### 3.6 Simulations for Test Data

Ideas presented above are checked on data from different distributions below. The reader is reminded that the method presented is non-parametric and thus will be expected to perform well on data it knows nothing about, so the goal of this is to determine how well the method performs for known distributions. The authors of [1] have specified that the method works best for a pdf bounded away from zero and infinity, but it may be useful to investigate how it performs for some other pdfs so that a decision to utilise this method for such will recognise the pros and the cons attached with a prior expectedly sub-optimal method. For each, the pdf is first estimated and then the entropy estimated, the same is repeated for anomalous situations, asymptotic results tested and anomaly detection capability shown. Also issues like changes in the pdf estimate with variations in $k$ will be considered for each one.

### 3.6.1 One Dimensional Gaussian Distribution

The pdf of a univariate Gaussian distribution is given by:

$$f(x) = \frac{1}{\sigma\sqrt{2\pi}} e^{-\frac{(x-\mu)^2}{2\sigma^2}} \qquad (10)$$

Where $\mu$ is the mean and $\sigma^2$ is the variance. These two parameters completely define the distribution.

The pdf of this distribution for ($\mu = 0$, $\sigma^2 = 2$) is shown in Figure 3.2 below. The task is to estimate this pdf using the k-nearest neighbour method described above. Realisations of data from the distribution was generated in matlab using the in-built randn(.) function. The code can be found in the appendix.

The estimated pdf is shown for $M = 2500$ and $k = 50$ which is the square root of $M$. Furthermore, it is important to note the differences in the figures that follow, which are results of varying values of $k$ and $M$. No conclusions will be drawn at this stage, only key observations are highlighted. It should however be noted here that except otherwise stated, choices of numbers are arbitrary.

Figure 3.3 is probably a very good estimate of Figure 3.2 especially when you consider that nothing about the dataset needs to be known to achieve this.
A major requirement of equation (5) is that $\hat{f}(x)$ should integrate to one. This ensures that the entropy estimate is correct. But knowing that we have access to only a discrete set, the numerical integration using rectangular area under the curve will converge to the same result for the continuous data set with more and more reduction in the width of the rectangle used.

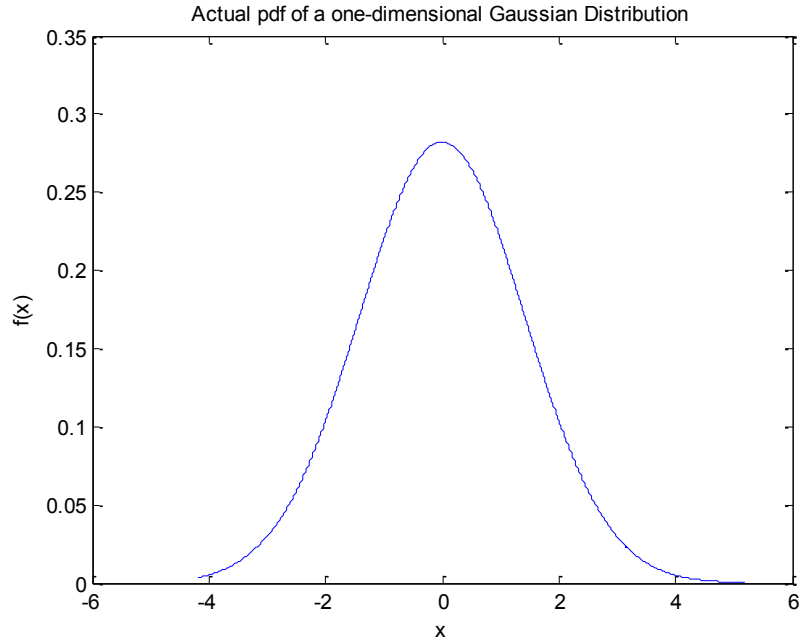

***Figure 3.2*** *Theoretical pdf of one-dimensional Gaussian with $\mu = 0$ and $\sigma^2 = 2$.*

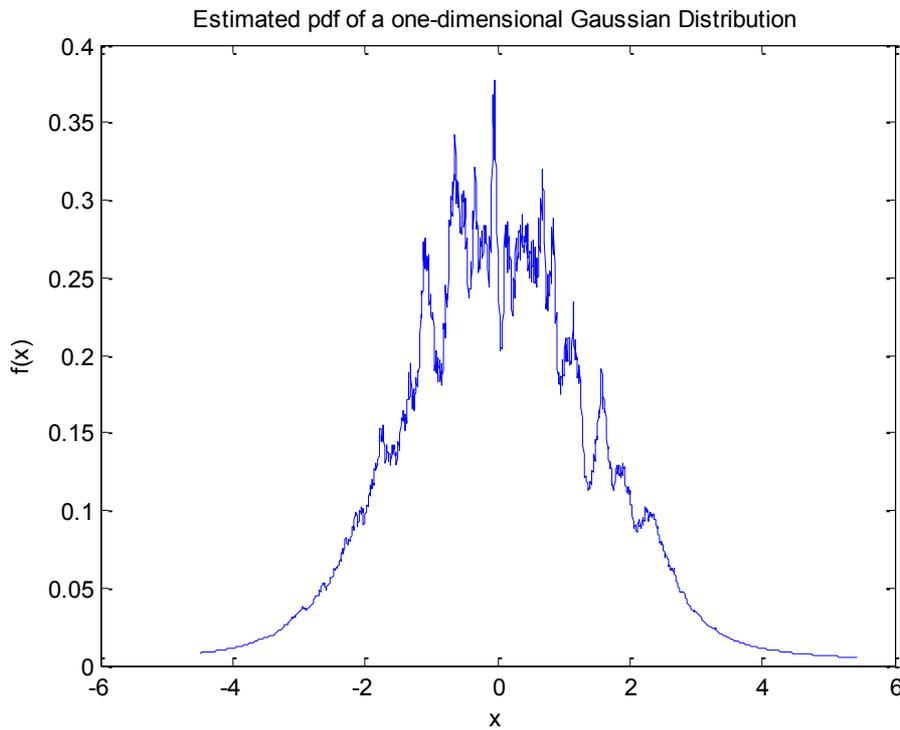

***Figure 3.3*** *Estimated pdf of one-dimensional Gaussian with $\mu = 0$ and $\sigma^2 = 2$.*

The rule of thumb is usually to use a width much smaller than the smallest separation between data samples, use a uniform grid, and use as many points as possible. This is reasonable because it prevents the overlapping of the rectangles and removes the errors that occur as a result of that, but at this stage nothing can be said of how this will perform for data in higher dimensions. It is also important to note the limitations of the simulation software to be used. Matlab is limited and cannot allow for the fulfilment of this condition for a large dataset with small variance. This is the major limitation of this approach. But theoretically speaking, it is a more accurate method.

A critical analysis of Figure 3.4 shows that for *(a).* $k = 10$, *(b).* $k = 20$, the estimates are quite poor. But for *(c)* $k = 75$, *(d)* $k = 100$, the estimates are better than Figure 3.3 with estimate becoming less 'noisy' and smoother, the question however is 'by how much?', and then the decision on the best value of $k$ would have to consider the computational complexity of achieving this also. The final choice depends on the peculiarities of the circumstances.

The q-q plot shown in the Figure 3.5 below compares the independent realisations of the normalised Shannon entropy estimate with the standard normal distribution. This verifies equation (9). Most of the samples lie along the straight line.

Finally confidence intervals are predicted for the Shannon entropy for increasing size of dataset in Figure 3.6 and it can be seen that this interval reduces with increase in data size as the estimates converge still with some bias compared to the true value. The true value is calculated from the closed form solution given by;

$$H = \frac{1}{2}\log 2\pi e\sigma^2 \qquad (11)$$

Where the variables retain their usual meanings.

Theoretically this closed form value will be reached as the dataset size continues to grow.

This is shown in blue in the plot of Figure 3.6.

Another thing to observe in Figure 3.6 is the reduction in variance of the estimator as the sample size increases.

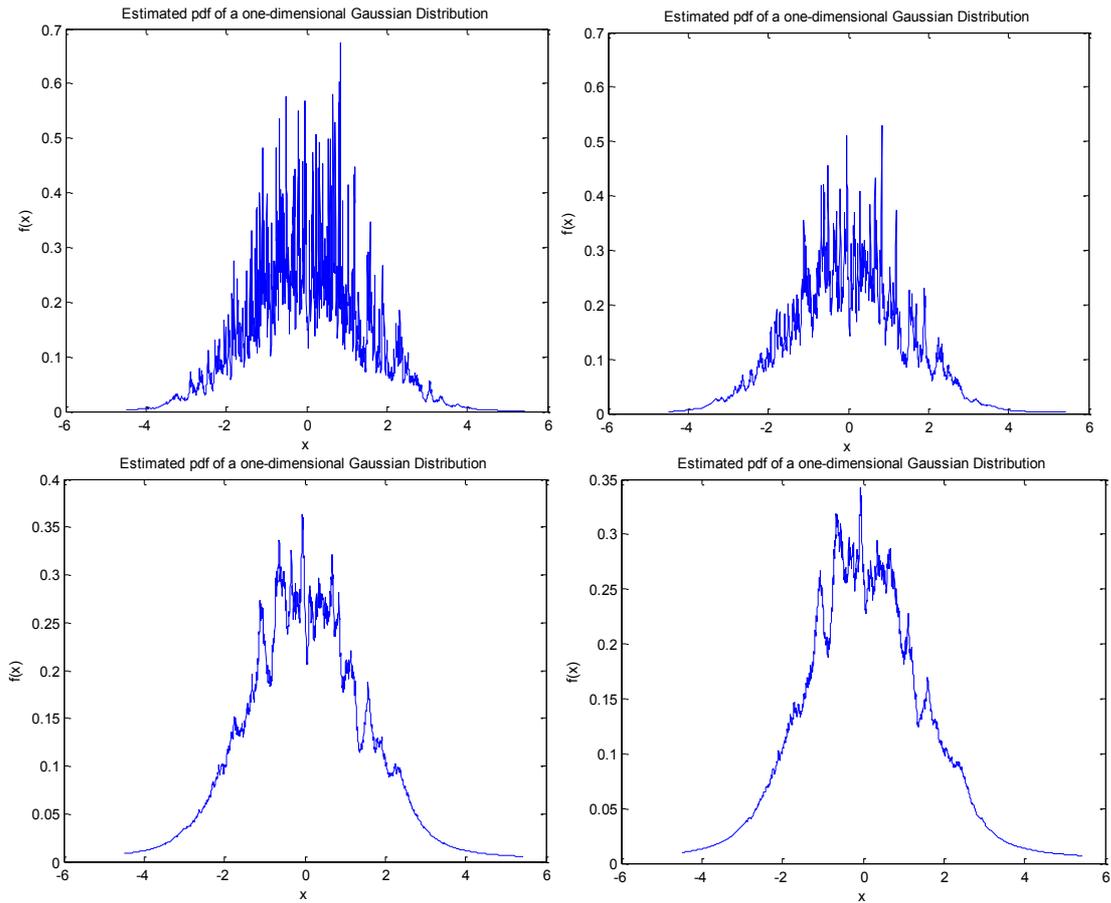

***Figure 3.4*** *Estimated pdf of one-dimensional Gaussian with μ = 0 and σ² = 2 for increasing values of k (a). k = 10, (b). k = 20, (c). k = 75, (d). k = 100.*

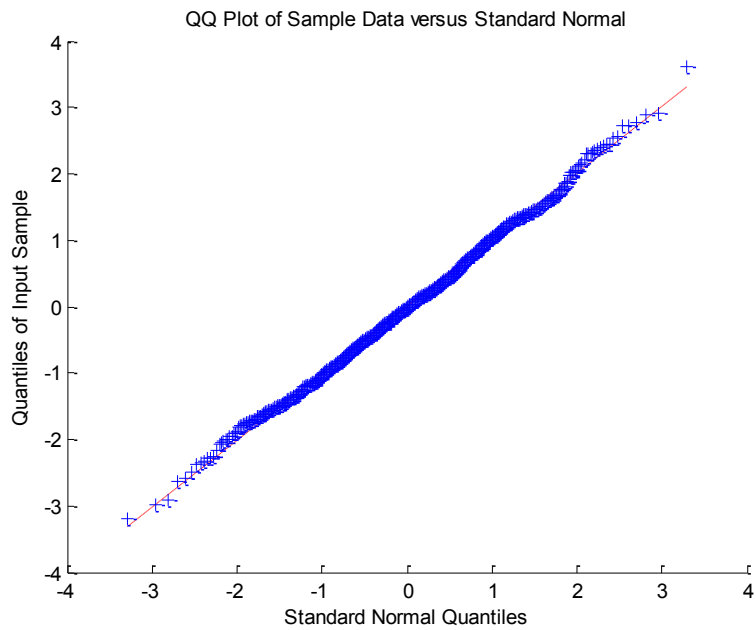

***Figure 3.5*** *q-q plot of the Shannon entropy estimate.*

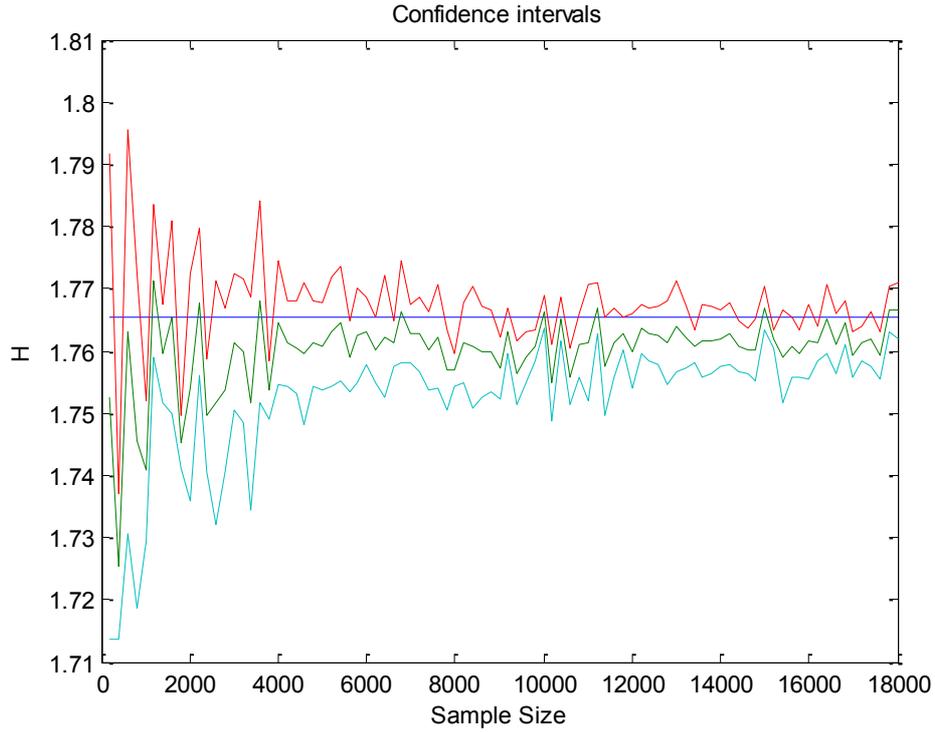

***Figure 3.6*** *95% confidence intervals on the Shannon entropy estimate for increasing sample size.*

Now to verify the method on anomaly detection, the dataset is intentionally altered to introduce anomalies at known points. The pdf of the new dataset is estimated as well as its entropy. Thresholds were set as follows;

$$t_\propto = \mu + z_{\alpha/2}\sigma \qquad (12)$$

Where $t_\propto$ represents the threshold, $z_{\alpha/2}$ represents the Z-score, with mean ($\mu$) and standard deviation ($\sigma$) of the estimates of the entropy for the normal data time points. Anomalies were introduced for designated time points. Mean and variance of the estimates of the entropy for the normal periods were used to set threshold for the anomalous point detection. The anomalies need to be quite large to be detected. This is intuitive because the method involves averaging the entropy values estimated from the pdf estimates for a large dataset. The larger the dataset the more the effect of the anomalies be reduced until it becomes nullified. Thus a large number of anomalies would be required for the method to detect them, or fewer but much larger anomalies may provide desirable results. But as shown in the ROC plot below, with some anomalies, the system seems to be able to achieve 0.9 detection rate for a false alarm rate of just less than 0.3. This is not too bad for a prior expectedly sub-optimal method for this kind of pdf. But the anomalies have to be large.

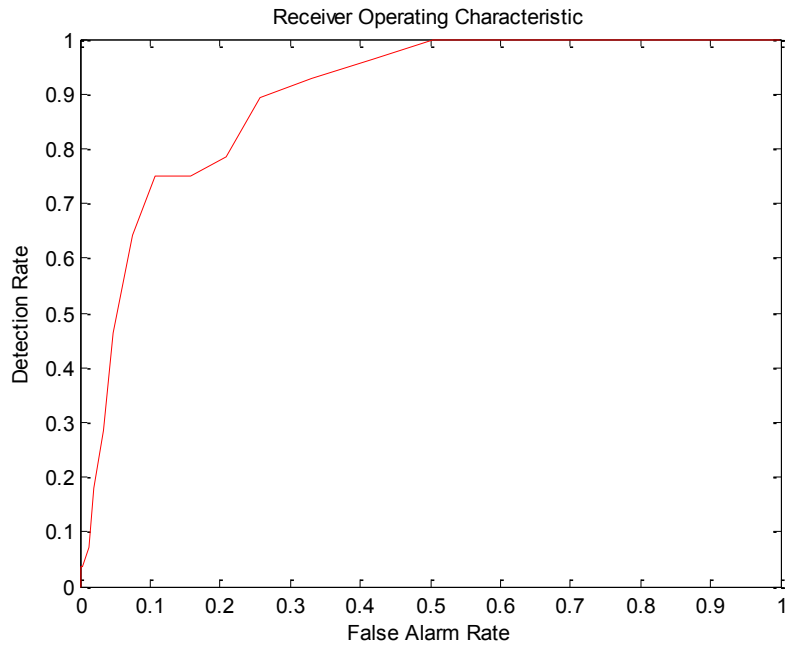

***Figure 3.7*** *Result of anomaly detection simulation on test data.*

The reader may want to take another look at Figure 3.3 to understand the difference between the normal condition and the anomalous condition of Figure 3.8.

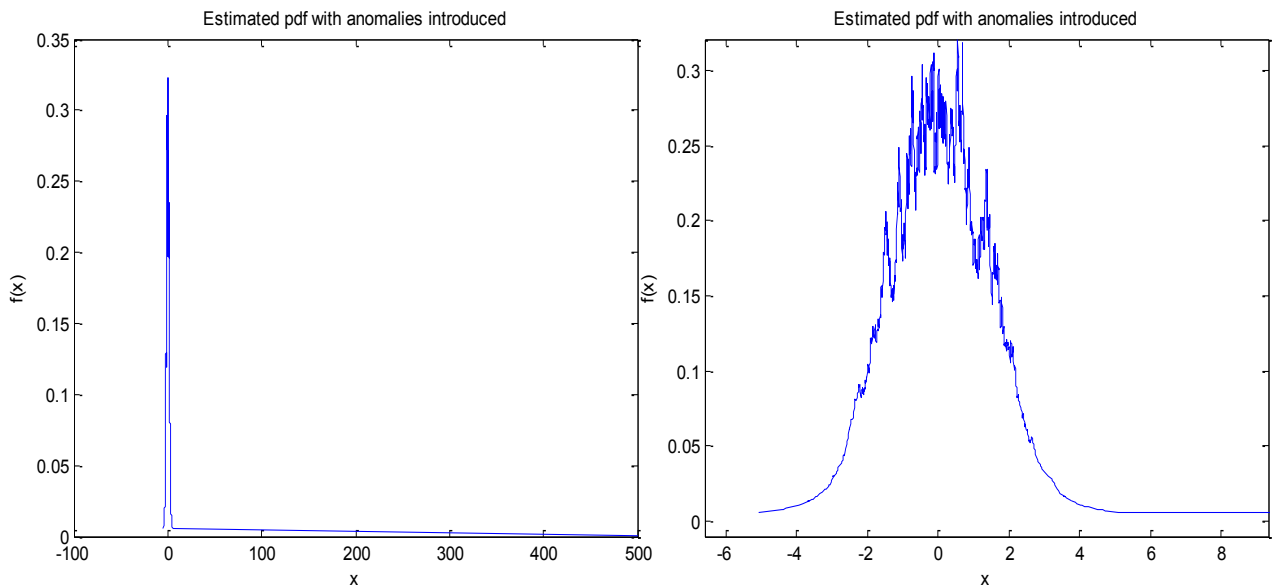

***Figure 3.8*** *Estimated pdf of the dataset with anomalies introduced (a). the actual figure. (b). a zoomed out version.*

A key point to look out for is that the pdf in Figure 3.8 has a long tail to the right (which may be on any side depending on what the values of the anomalies are in the situation), and the general shape and height of the plot is also quite different. It goes without saying from the result so far obtained that this is definitely not the best that can be done (basically because of the sheer largeness of the anomalies that need to be present), but then, it is too early to conclude.

### 3.6.2 Two Dimensional Gaussian Distribution

This is pretty much the same as before except now the data is in two dimensions, therefore sorting can no longer be done like in the one dimensional case and yes, scalability issues begin to surface. A brief description of how the simulation is carried out leaves questions about how easy it would be to analyse a dataset of higher dimensions.

Firstly, the dataset is divided into and the first part is used to estimate the pdf as before. The Euclidean distance in two dimensions has to be calculated while also the k-nn region is now circular as opposed to rectangular in the previous case and the volume is updated to reflect that.

Secondly, issues with the pdf integrating to one urges one to go in the direction of infinite data, but worse still, two dimensions have to be considered and with the pdf, the plot is definitely going to be three dimensional. To achieve this, the meshgrid(.) function is used, which takes a very long time to run and throws up the 'out of memory' error even for a dataset that should be considered as not so large for two dimensions because for an arbitrary $N$, the actual pdf calculations will be done for $N^2$ number of points. So theoretically, yes, the result of the pdf estimate will be accurate. But experimentally, one may have to work within the limits that the simulation software allows.

Apart from these, other steps are essentially the same. The actual pdf of the two dimensional Gaussian distribution is shown in Figure 3.9 for the given parameters. The estimated pdf is then shown in Figure 3.10, this is quite close to the actual pdf shown in Figure 3.9.

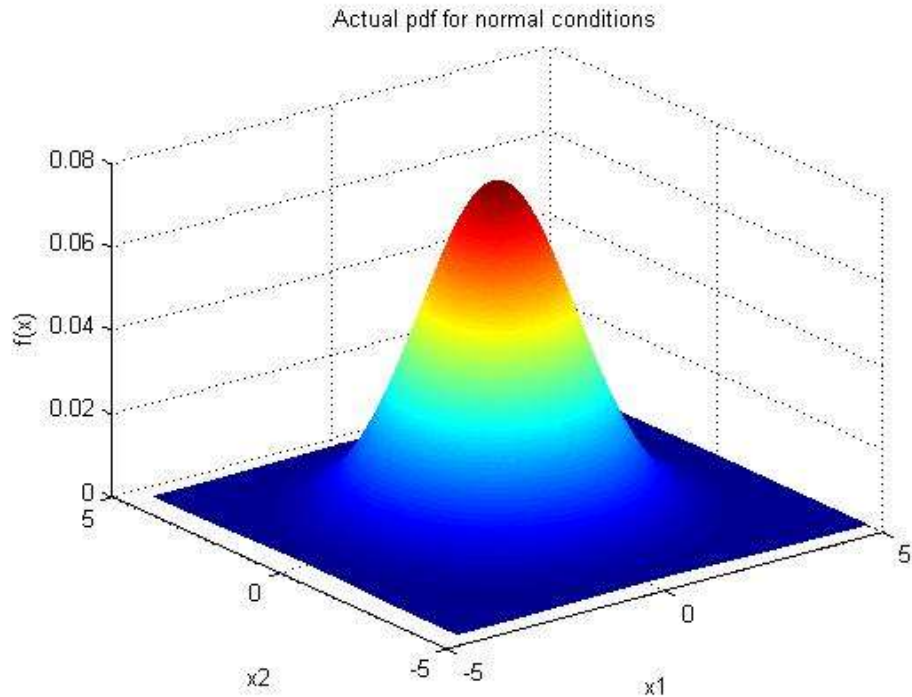

***Figure 3.9*** *Actual pdf of two-dimensional Gaussian with μ = 0 and σ² = 2.*

Changes in the pdf estimate with varying $k$ is shown in Figure 3.11. The main ideas are not different from those highlighted in the previous section; the pdf becomes smoother as the $k$ increases.

It is worthy to note here that there is no issue with the boundary for Gaussian data, therefore boundary correction is not needed.

Figure 3.12 show that the estimate of the entropy for two dimensional Gaussian data is poor at the beginning with very high bias, but this reduces quite quickly as the estimate appears to be quite sensitive to increase in data size and from a data size of 10000, the estimate is much better. The variance also seems to reduce.

The poor estimate obtained with quite a small dataset is thought to be due to issues with the curse of dimensionality i.e. more data points are required for the estimate to be near correct in two dimensions than in one dimension; the scalability of the method as the dimension increase and also because the pdf is not bounded. This highlights the requirement in [1] that the pdf be bounded away from zero and infinity.

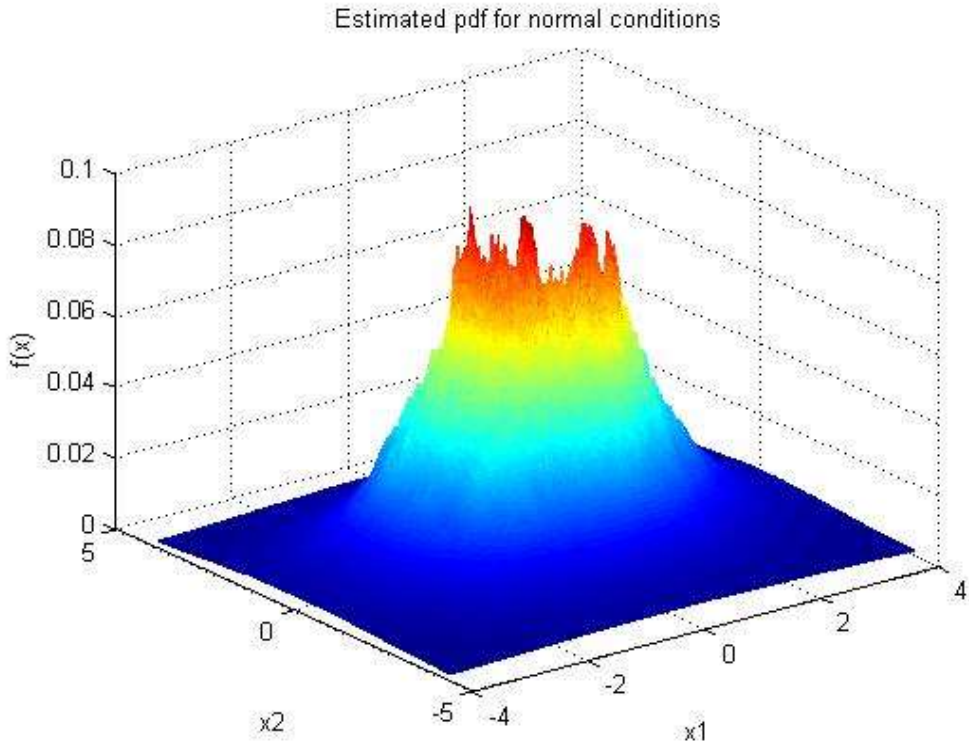

***Figure 3.10*** *Estimated pdf of two-dimensional Gaussian with $\mu = 0$ and $\sigma^2 = 2$.*

It is therefore intuitive to think that this method will not perform quite well with the two dimensional Gaussian as even the q-q plot only weakly shows a convergence to the standard normal distribution as shown in Figure 3.13 below.

How will this system perform when anomalies are introduced? The intuitive answer is 'less than obtained for the one dimensional form'. Anomalies were introduced at known points and the pdf of the anomalous dataset was estimated. The entropy was then estimated and the result obtained is presented.

Firstly, the pdf of the anomalous dataset is shown in Figure 3.14. The pdf now stretches to well beyond the normal dataset restrictions.

The ROC plot in Figure 3.15 confirms the fears. It only reaches a detection rate of 0.6 for a huge false alarm rate of 0.5! The method performs poorly for the two-dimensional Gaussian. This also suggests that maybe the effect was only suppressed for the one-dimensional form, and further increase in dimensions may see worse performance.

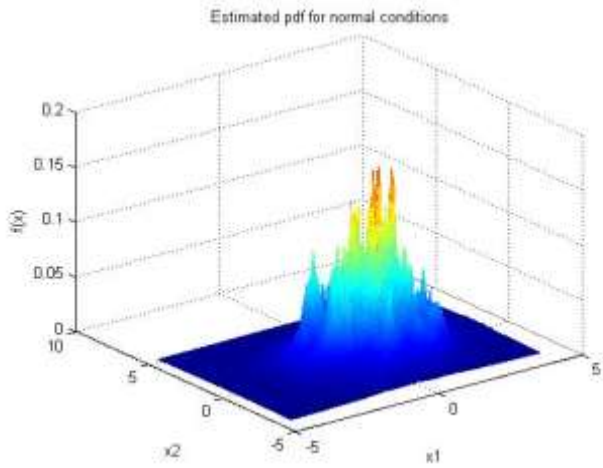 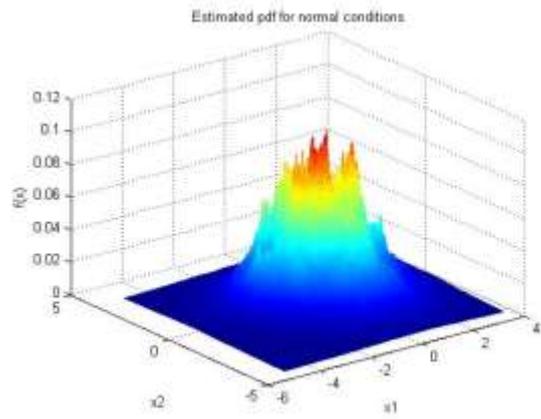

*(a).*  *(b).*

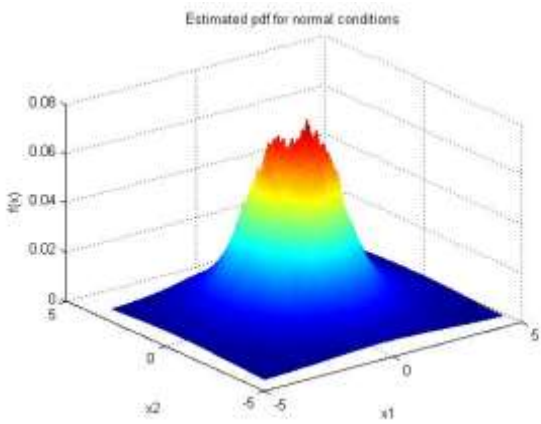 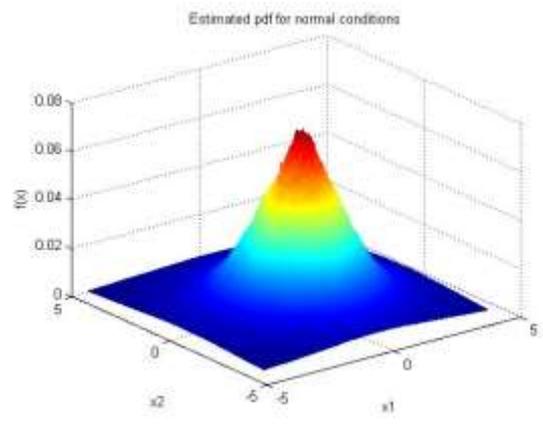

*(c).*  *(d).*

***Figure 3.11*** *Estimated pdf of two-dimensional Gaussian with $\mu = 0$ and $\sigma^2 = 2$ for*
*(a) $k = 10$, (b) $k = 20$, (c). $k = 75$, (d). $k = 100$.*

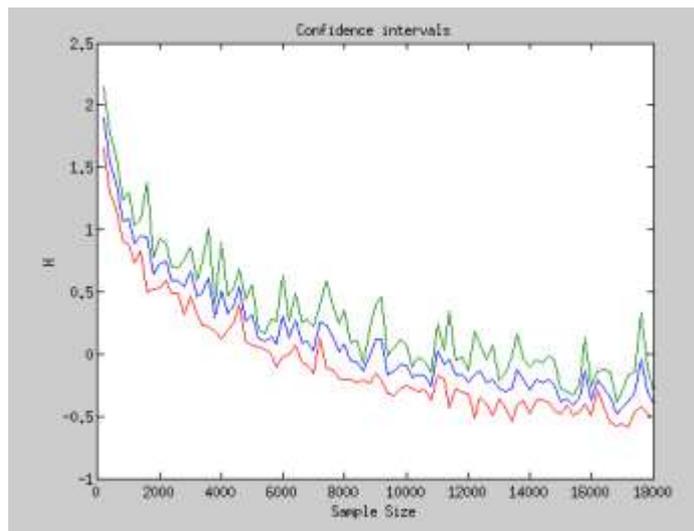

***Figure 3.12*** *confidence intervals on the Shannon entropy estimate for increasing sample size.*

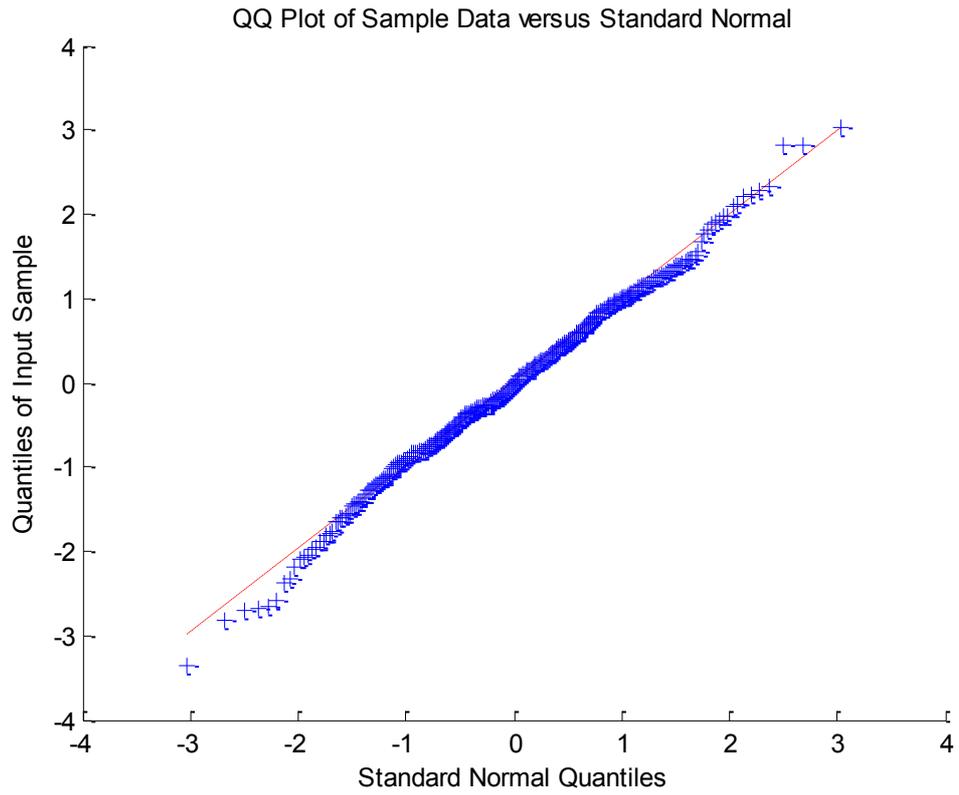

*Figure 3.13* *q-q plot of the Shannon entropy estimate.*

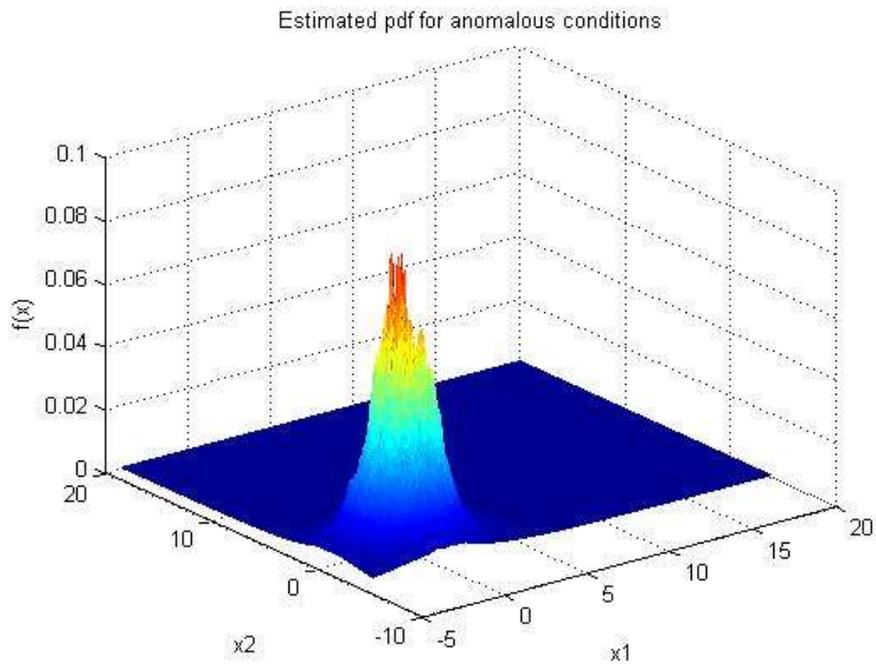

*Figure 3.14* *Estimated pdf for anomalous two-dimensional Gaussian.*

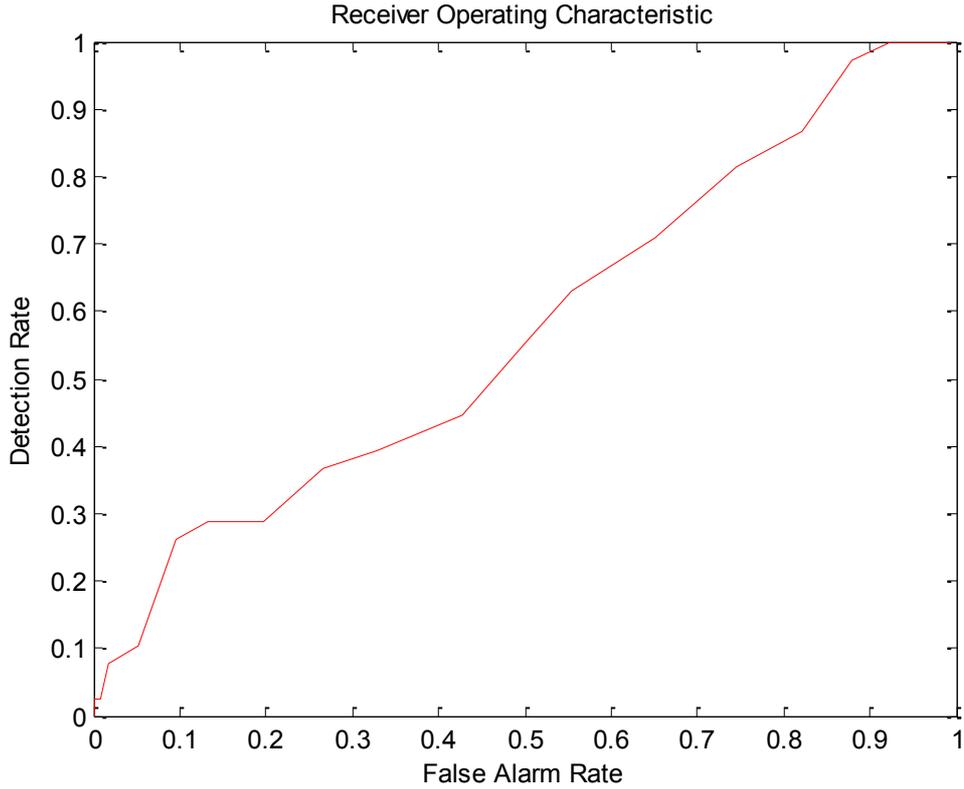

*Figure 3.15 ROC plot for two-dimensional Gaussian*

### 3.6.3 One Dimensional Beta Distribution

The beta distribution is the next type of distribution where the pdf estimation method will be tested. The beta distribution with parameters ($\alpha = \beta = 4$) are used for this work. This pdf is very similar to the Gaussian as will be seen but it is bounded away from $\pm\infty$ unlike the Gaussian case. This raises expectations for both one-dimensional and two-dimensional forms.

The pdf of the beta distribution is given by:

$$f(x) = \frac{x^{\alpha-1}(1-x)^{\beta-1}}{B(\alpha,\beta)} \qquad (13)$$

Where $\alpha, \beta$ are the parameters. In this case, $\alpha = \beta = 4$.

This pdf is shown in Figure 3.16. Independent realisations of beta distributed random variables are generated using betarnd(.), and the method so far described was applied to the data to estimate its pdf. The result obtained is shown in Figure 3.17. How close can one get to Figure 3.16? Or better still how close is close enough? This is application dependent as well system dependent. Differences can again be seen in the plots for the different estimates for varying $k$.

The results are as expected.

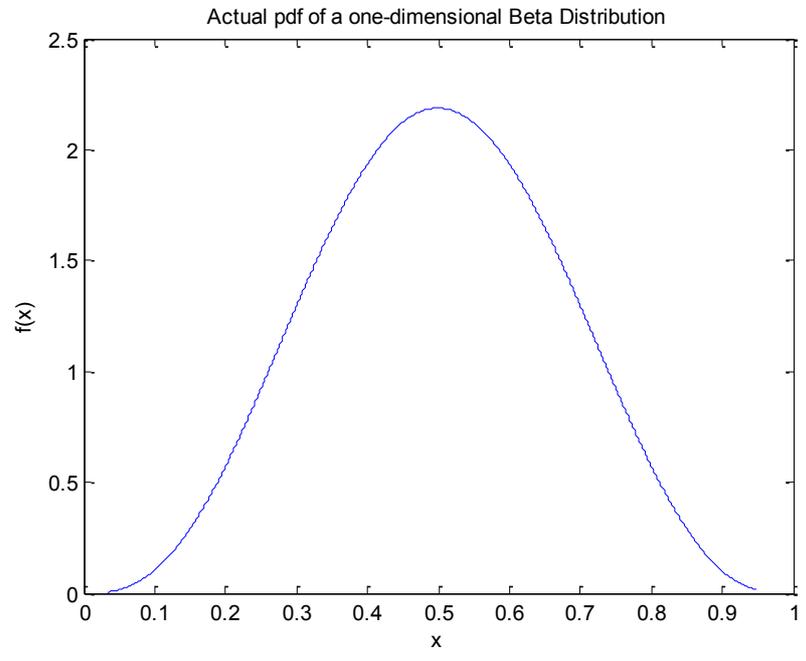

*Figure 3.16* Theoretical pdf of one-dimensional Beta Distribution.

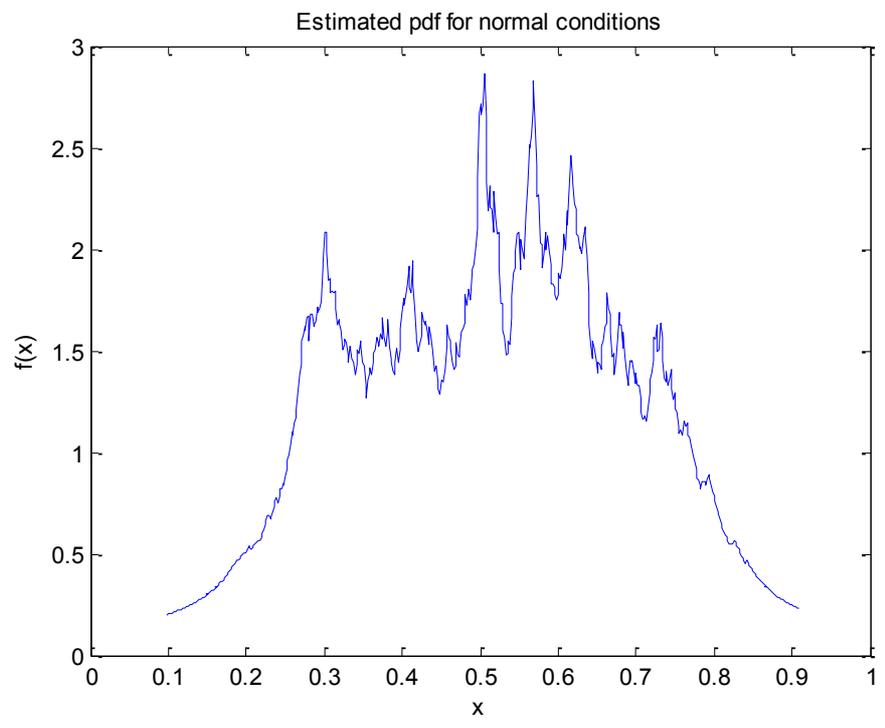

*Figure 3.17* Estimated pdf of one-dimensional Beta Distribution.

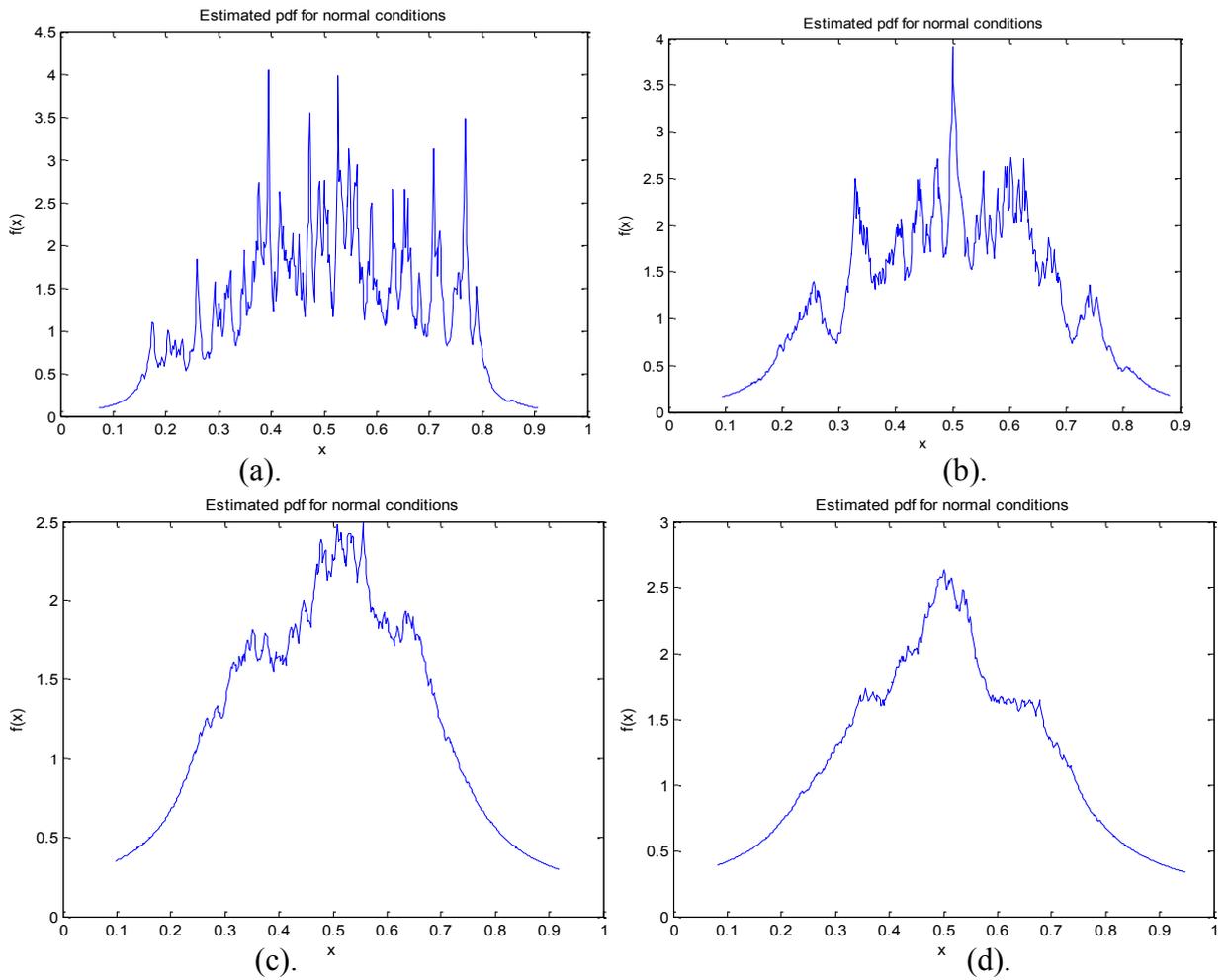

***Figure 3.18*** *Estimated pdf of one-dimensional Beta Distribution* for increasing values of k (a). $k = 10$, (b). $k = 20$, (c). $k = 75$, (d). $k = 100$.

When anomalies are introduced, the pdf estimate is shown in Figure 3.19 below.

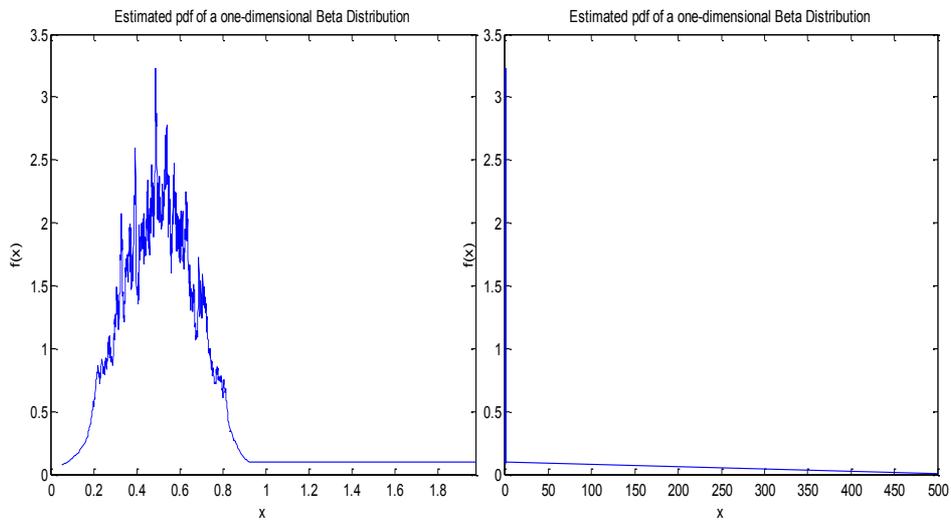

***Figure 3.19*** *Estimated pdf of one-dimensional Beta Distribution* with anomalies.

As in other cases, the q-q plot of the normalised Shannon entropy estimate is shown in the Figure 3.20 below. And as earlier suspected, the ROC plot for the beta distribution performs much better than the Gaussian.

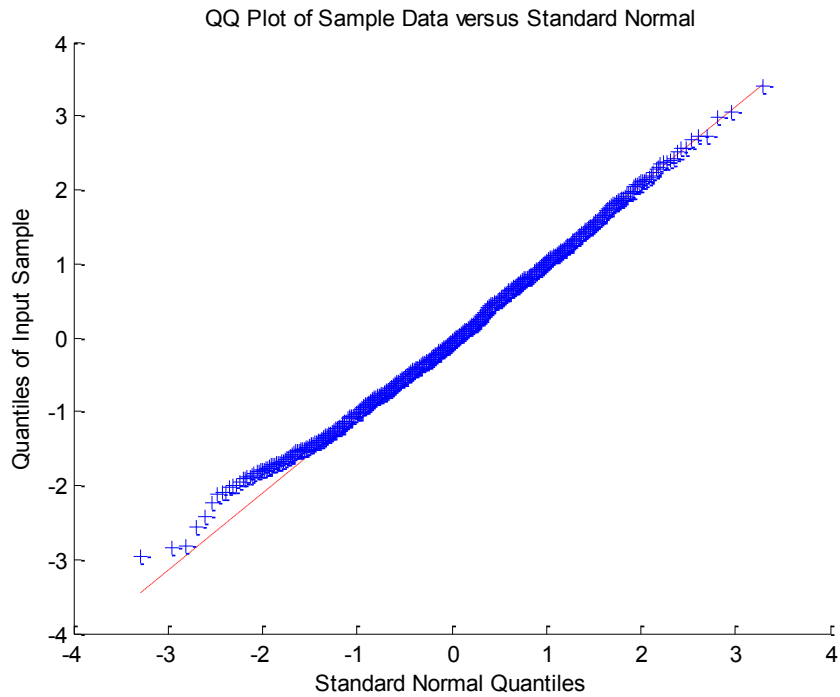

*Figure 3.20 q-q plot of the Shannon entropy estimate.*

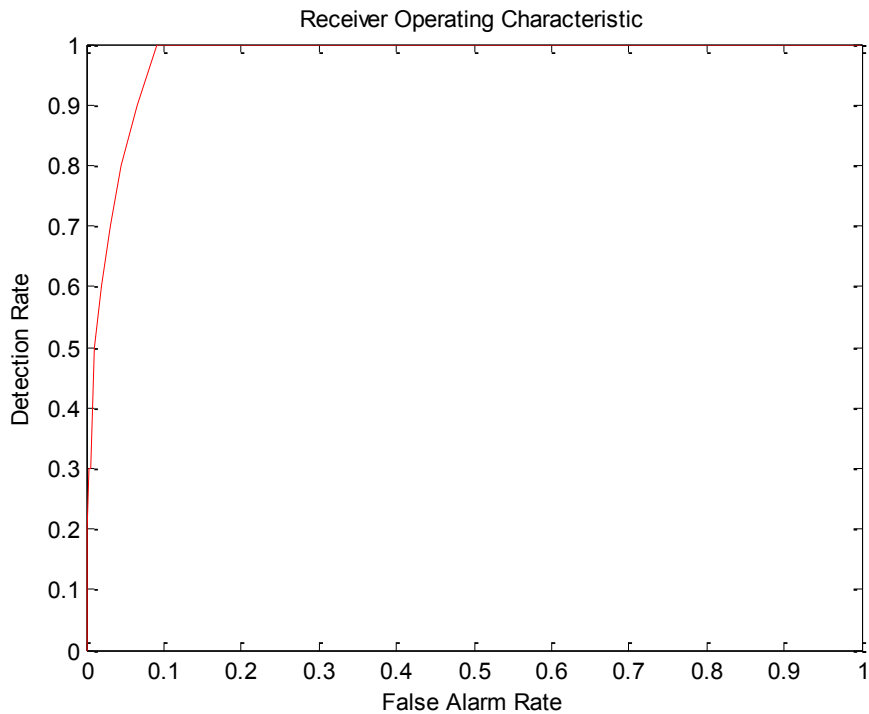

*Figure 3.21 ROC plot for one-dimensional Beta Distribution*

It attains a detection rate of over 0.95 for a false alarm rate of 0.05.

### 3.6.4 Two Dimensional Beta Distribution

The two dimensional beta distribution is shown in Figure 3.22 below. Independent realisations were generated in matlab as in the last section only that this is in two dimensions and some ideas from the two dimensional Gaussian pass on directly here. The pdf estimate for the normal condition is also shown in the Figure 3.23.

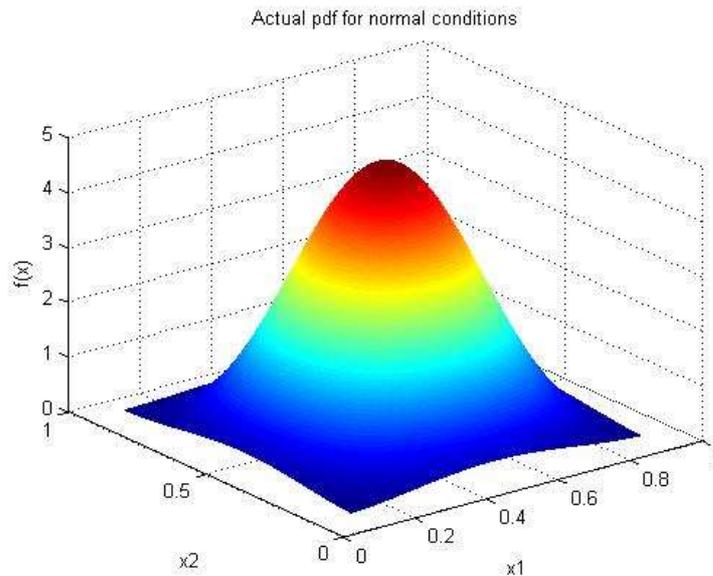

*Figure 3.22* *Actual pdf of two-dimensional Beta Distribution*

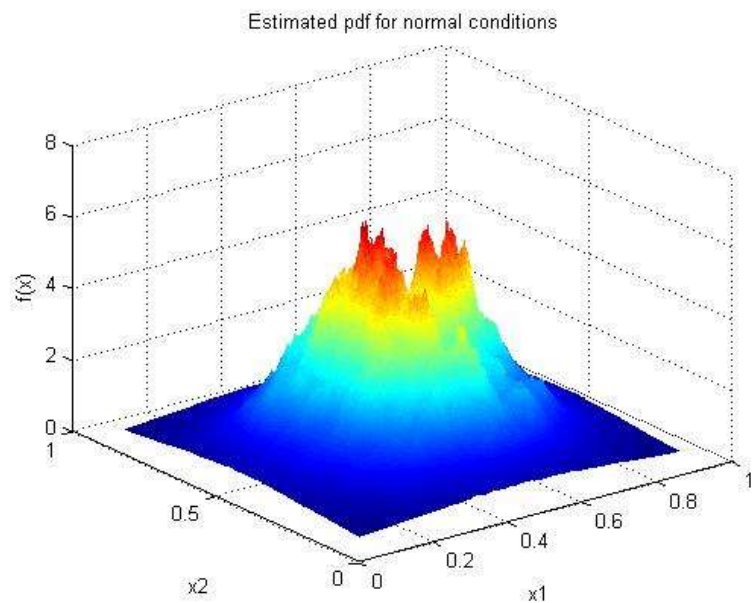

*Figure 3.23* *Estimated pdf of two-dimensional Beta Distribution*

The argument about the effect of the changes in $k$ continues on the same path here. The result is no different from the ones previously obtained. Changes in the estimate for increasing values of $k$ is shown.

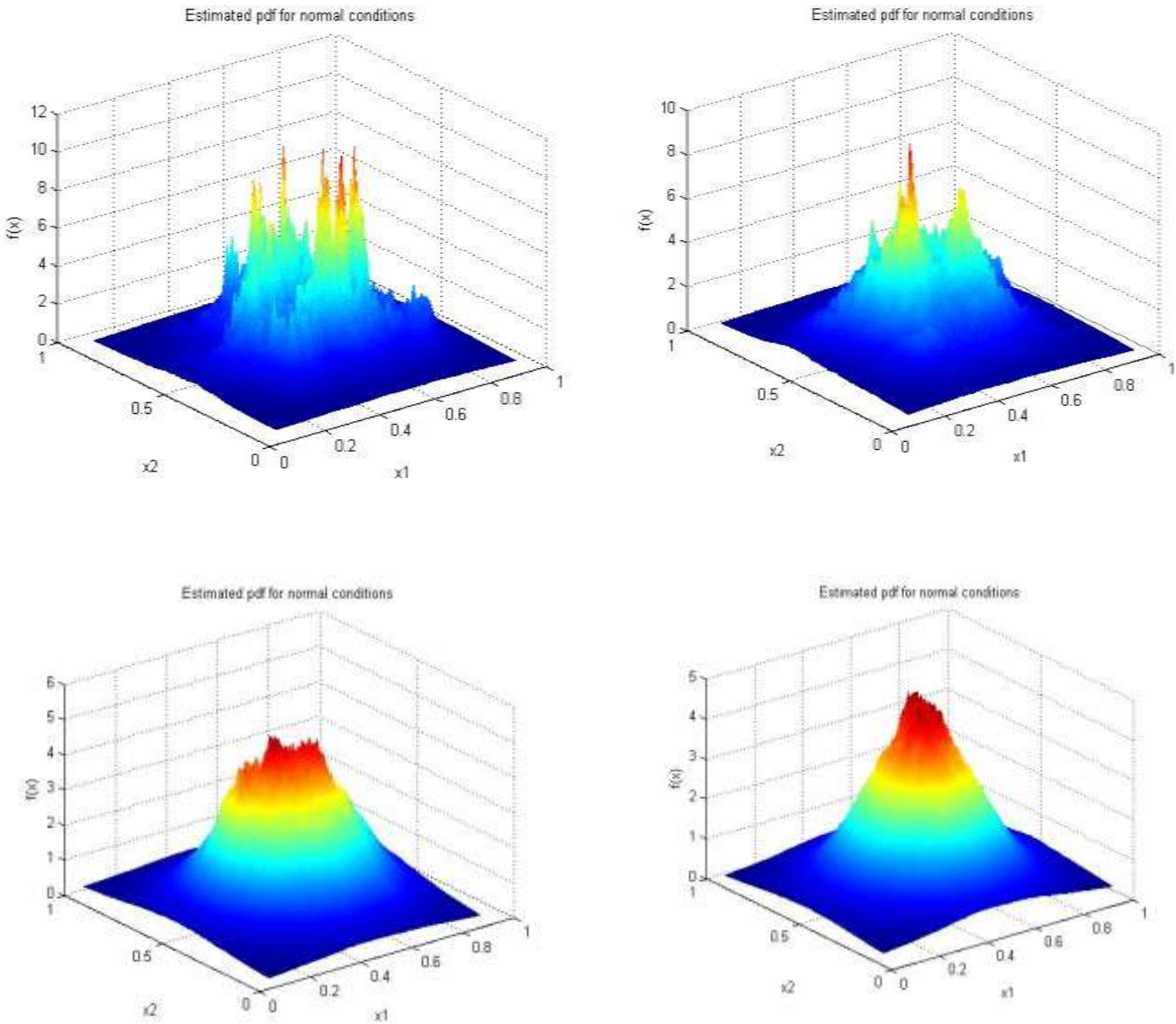

***Figure 3.24** Estimated pdf of two-dimensional Beta Distribution* for increasing values of $k$

The estimate of the two-dimensional beta distribution dataset with anomalous data is shown in Figure 3.25 below.

The 95% confidence intervals on the entropy estimates with increasing sample size looks better than that for the two dimensional Gaussian in the sense that a threshold can be more easily placed in Figure 3.26 that would give a result with much less false alarm. This would be understood if one realises that the threshold that would be required for a data sample size of 4000 in Figure 3.12, is much different from say, size of 8000. This leads to being on two extremes, either one accepts a sub-optimal system with very high missed detection, or one accepts the high false alarm rates. The downside of that is that the detector would get ignored by would-be respondents. There would be much less false alarms and missed detection in the former than in the latter. The true value in this case is -0.75.

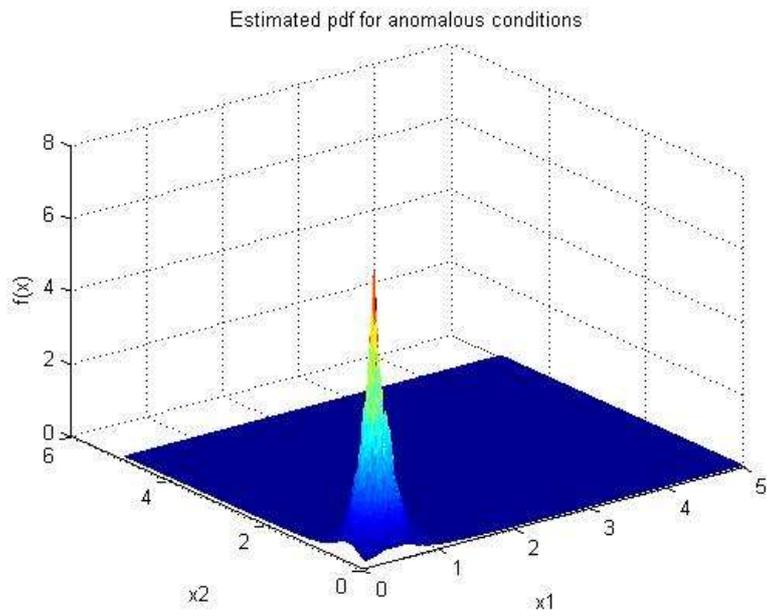

*Figure 3.25* *Estimated pdf of two-dimensional Beta Distribution* with anomalies.

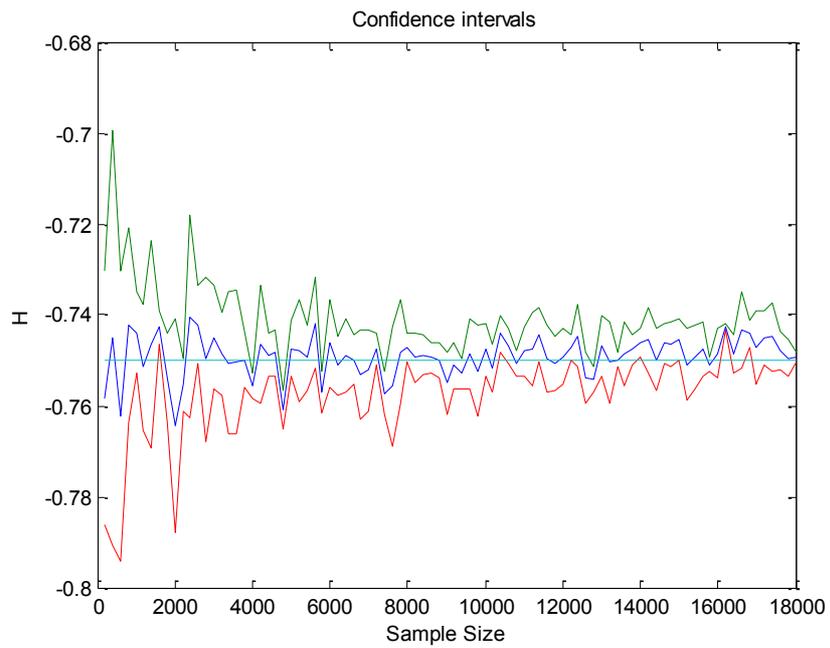

*Figure 3.26* *confidence intervals on the Shannon entropy estimate*

The q-q plot also confirms the asymptotic normality of the normalised Shannon entropy estimator for two dimensional Beta distribution. This is shown in Figure 3.27 below.

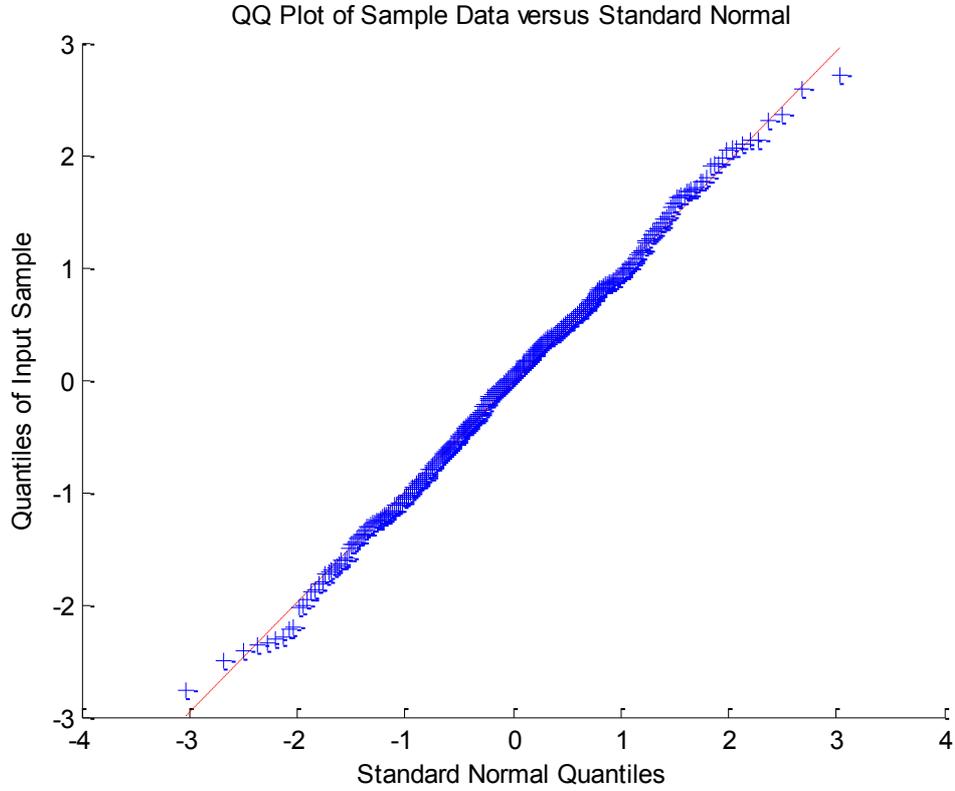

*Figure 3.27 q-q plot of the Shannon entropy estimate.*

### 3.6.5 Two Dimensional Mixture Density (Beta and Uniform Distributions)

This is the specified distribution in the paper where the method has been tested and verified to work by the authors of [1]. The results from [1] concerning this will be verified.

The two dimensional mixture density is written as:
$$f_m = pf_b + (1-p)f_u \qquad (14)$$
Where: $f_b$ is the two dimensional beta distribution with parameters same as before, and $f_u$ is a uniform density and $p$ is the mixing ratio.

The first task here is to estimate this pdf. The actual pdf of this mixture distribution is shown in the Figure 3.28.

In order to estimate this pdf, the components of the mixture are first estimated and then added together in the following order; Upon generation of the data from the two dimensional beta distribution, the pdf is estimated and then the uniform density pdf is then added to arrive at the mixture estimate. This forms the basis on which other results are established.

The actual pdf shown in the figure below is that for the case where the data is independent in both directions. This is very much like the two dimensional beta distribution in shape. The only difference being that it has been bounded away from zero to 0.2, and from $+\infty$ to 1. And in fact, the beta distribution with parameters (4,4) is very much like the Gaussian. The reader may have observed this in the previous sections. The estimated pdf is shown in Figure 3.29. So what happens when anomalies are added? This is shown in Figure 3.30.

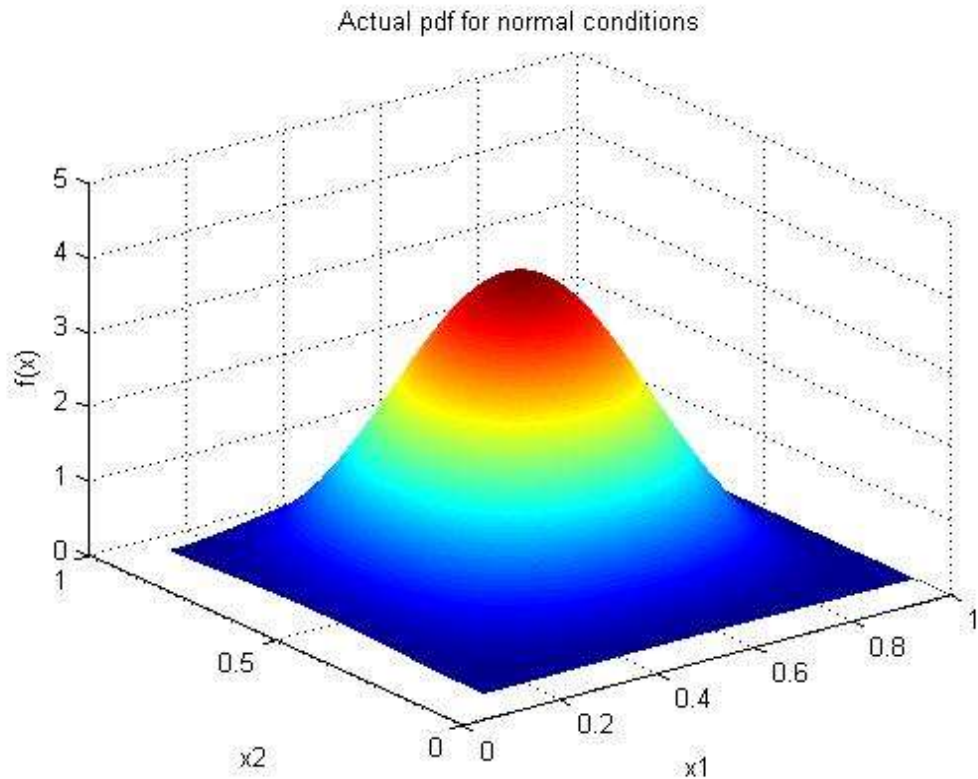

*Figure 3.28* *Actual pdf of two-dimensional Mixture Distribution*

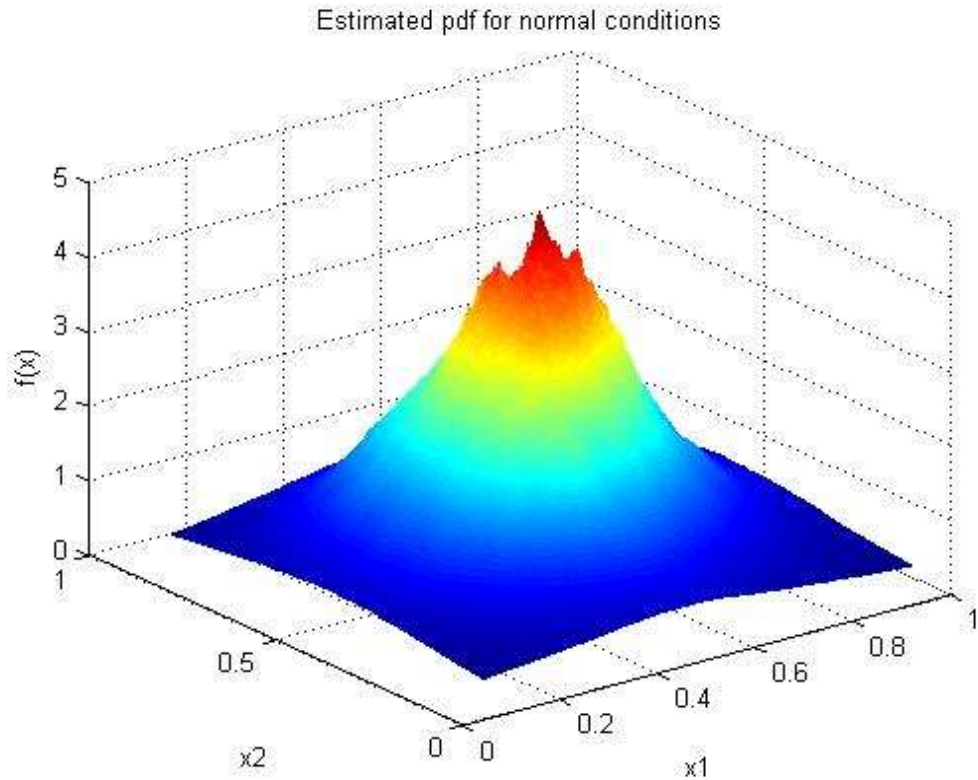

*Figure 3.29* *Estimated pdf of two-dimensional Mixture Distribution*

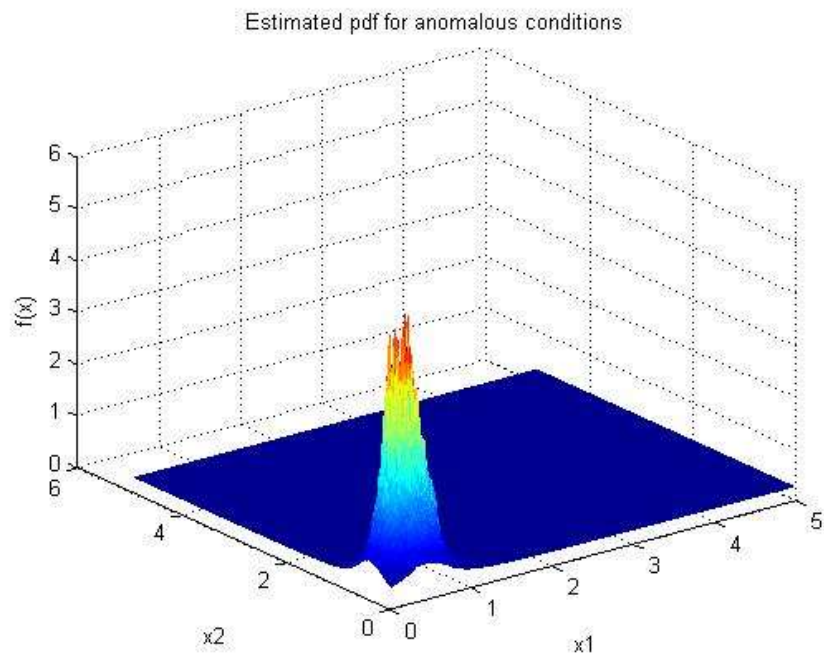

*Figure 3.30* *Estimated pdf of two-dimensional Mixture Distribution with anomalies*

The 95% confidence intervals are shown in Figure 3.31, which has values different from the one obtained in [1]. This is particularly interesting because the method used here has been verified especially for the Gaussian distribution, which has a closed form solution and was found to converge at the true value with very little bias for large enough dataset. How then will one defend this? Well but what is known about this particular mixture distribution is that it has its largest pdf values greater than 1, and not so much less than one for the others, remember the pdf has been bounded on both ends; this can be seen in both Figure 3.28 and Figure 3.29. Therefore, applying equation (2) to this without even actually doing the calculations, one could expect to have a negative result, and this is what is presented below. In fact, this issue appeared to have been dealt with in [9] which was written by the same authors where the result presented for the three-dimensional form of this same distribution is much more plausible compared with that in [1]. But at the end of the day, entropy is just a number, and it may be interesting to begin to ask if it would be possible for two different distributions to have exactly the same entropy? An intuitive answer will be; Yes. If this holds true, it must then constitute the weakness of this algorithm.

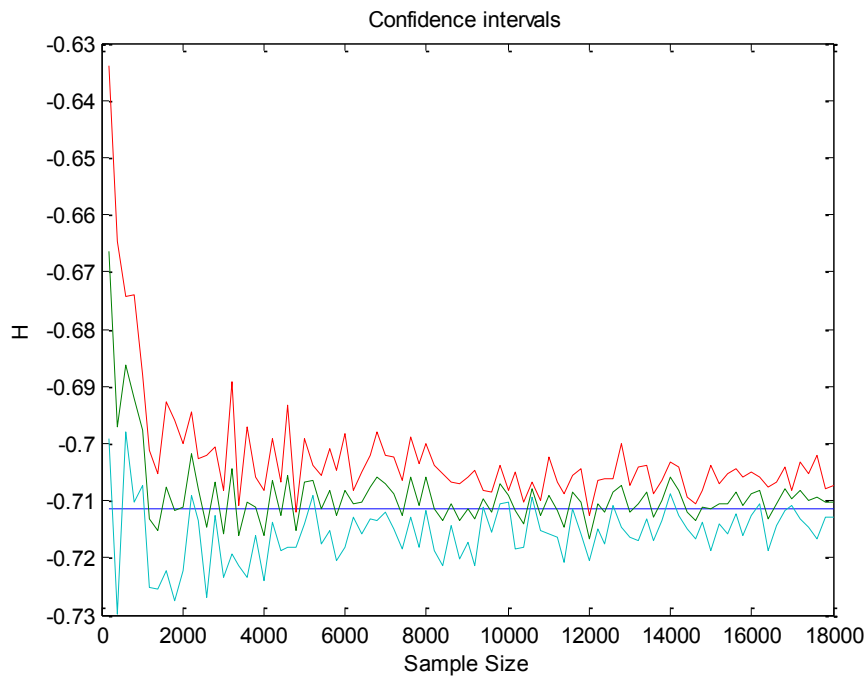

***Figure 3.31*** *confidence intervals on the Shannon entropy estimate*

The central ideas of the Figure 3.31 should however not be lost in spite of this. The figure shows that yes, the estimate has a high variance for smaller dataset sizes and this variance reduces as the size increases. The true value is –0.711.

The q-q plot is shown in Figure 3.32 for the normalised entropy, and it sure confirms the claims of the authors of [1].

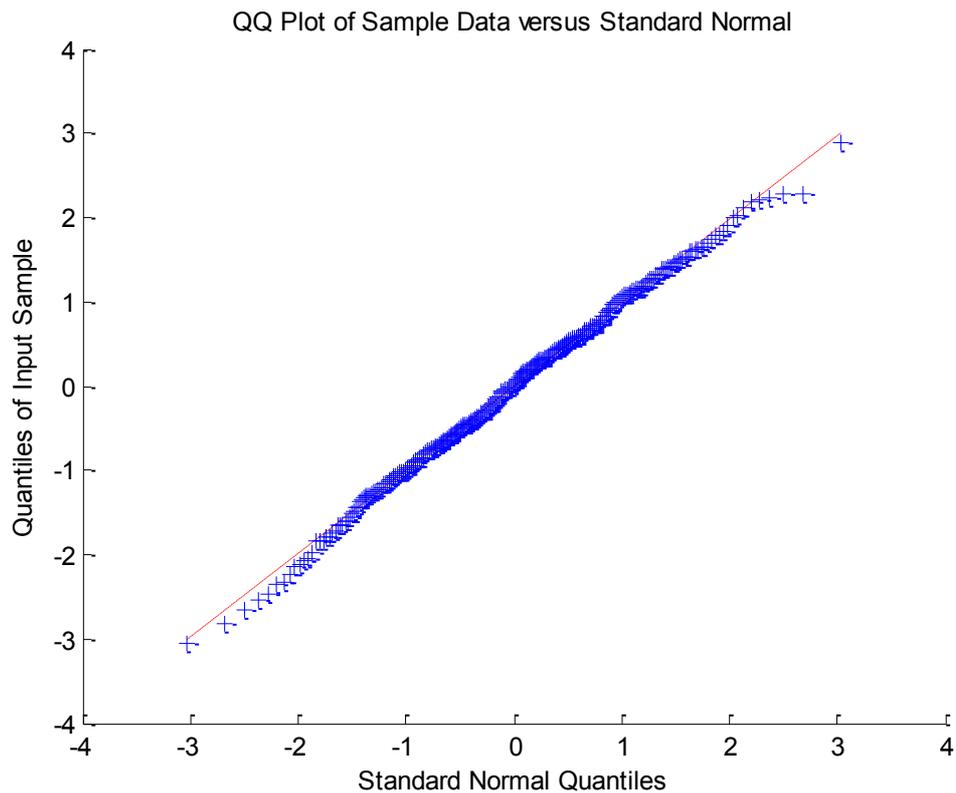

*Figure 3.32 q-q plot of the Shannon entropy estimate.*

### 3.7 Simulation on real data.
#### 3.7.1 Introduction

The data was obtained from an experiment carried out at the University of Michigan in 2006. [28] It involves sensors set up in a lab environment, which communicate with each other via broadcast and each sensor measures the received signal strength indicator (RSSI) from all other sensors. Measurements were taken over a period of 30 minutes. More information and raw data can be found at [28].

### 3.7.2 Main Experiment

Students were to come in and out of the lab at predefined times to obstruct sensor communications thus introducing anomalies which took the form of reduction in RSSI values. Also, there was a camera used to keep the ground truth. This is then compared with the results returned by the sensors and the anomaly detection method presented to determine how well it works.

### 3.7.3 Pre-processing

Because the measurement times were not at exactly the same time (i.e. they were not synchronous), there was need to pre-process data by interpolating. Also, to remove effects of temperature drifts, the local means were removed. Further information about this can be found at [28].

### 3.7.4 Detection

The dataset from time 1 -50 are specified as the normal conditions while data from outside this may contain the anomalous data to be detected. So the method of entropy estimation presented is calculated using the normal points, and are then used to detect anomalies in the anomalous points. The threshold is set from the values obtained for the normal condition as given in equation (1) using estimates of the mean and variance of the entropy estimate for each time point. The result presented below are for the temporal anomaly detection, where the algorithm was applied on data obtained for different times.

### 3.7.5 Results

By applying the method to this data, the following results were obtained. Firstly, the pdf estimate for normal condition is as shown in the Figure 3.33. Of course, there is no actual pdf here. This pdf estimate could be obtained from data for any of the times from 1-50. With the same reasoning, any anomalous time instant measurement could be used to estimate the pdf for an anomalous dataset. To be sure of this, the ground truth data may be used to know what time instant data is anomalous. This can then be used to estimate the pdf. This is shown in Figure 3.34.

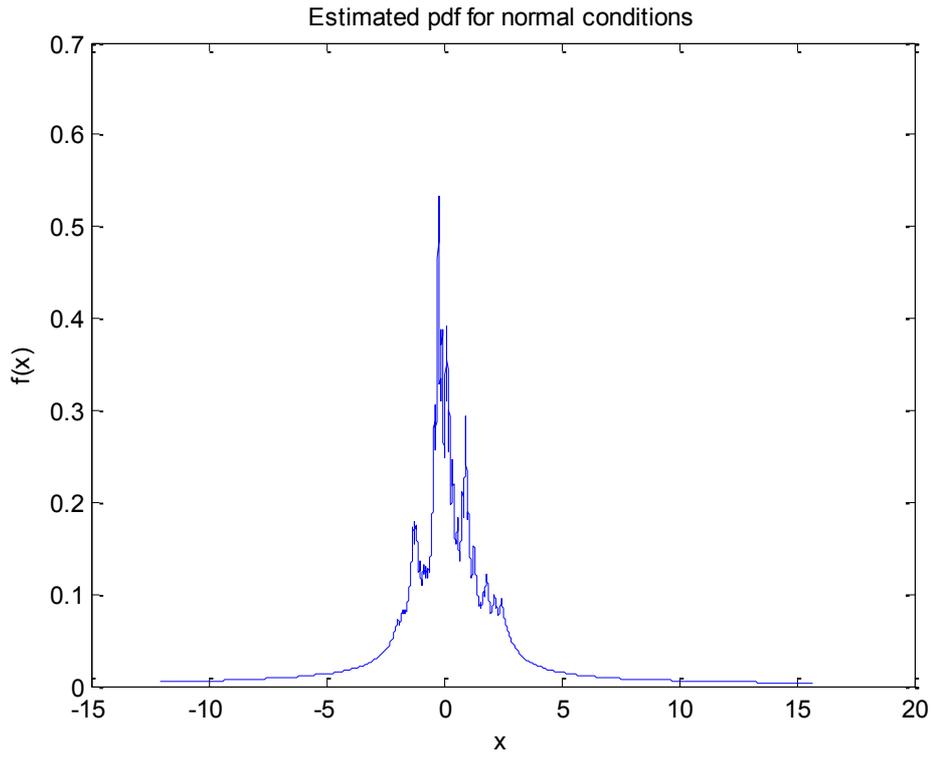

*Figure 3.33* Estimated pdf of the real data for normal condition

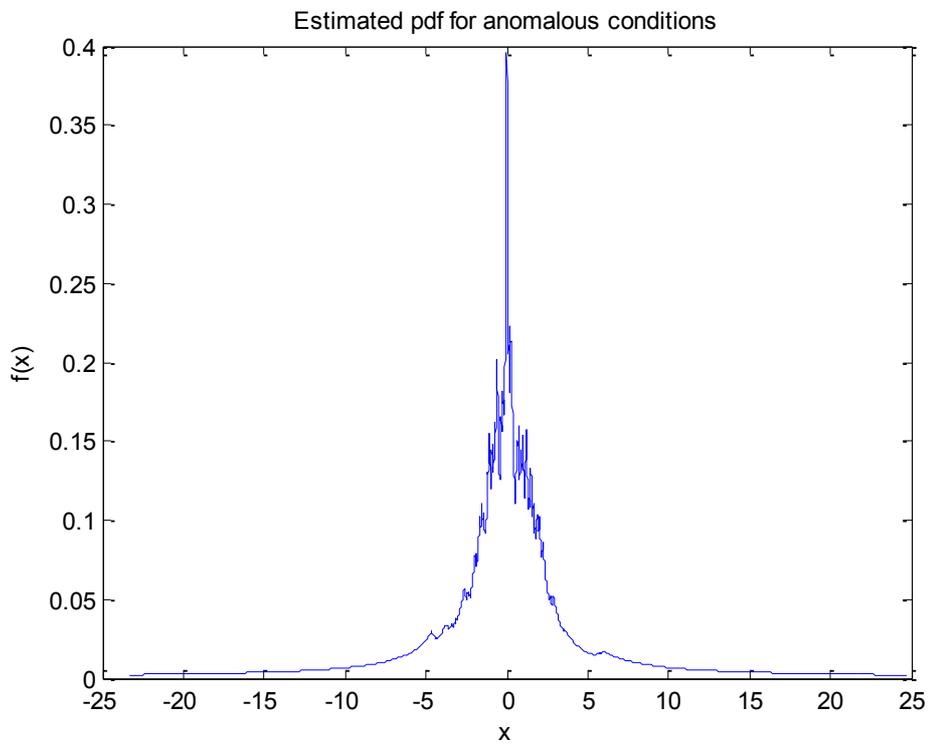

*Figure 3.34* Estimated pdf of the real data with anomalies

Estimates of the entropy were obtained using the method described and normalized. The result is then compared with the ground truth data, by plotting as a scan statistic in Figure 3.35 below. It is shown in the figure that the method performs pretty well on this kind of anomalies. The ground truth is shown in green while the entropy estimate is shown in blue.

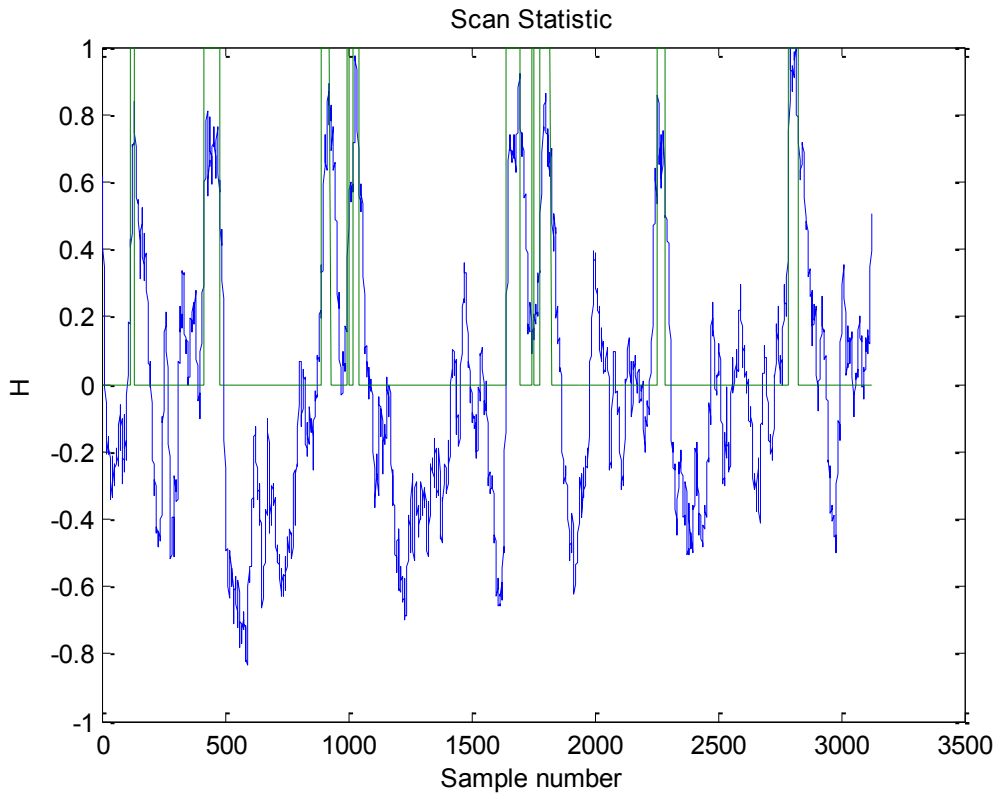

*Figure 3.35 Scan statistics for the entropy estimates for different time instants.*

The ROC plot for this is shown in Figure 3.36, which really looks promising and confirms the result in [1]. The only down point of this part is the q-q plot, which seems to be less optimal compared with all the others that have already been obtained and presented in this report. It is shown in Figure 3.37. Why does this happen? The quickest suspect is the dataset size. The method has been shown to provide entropy estimates that get better as the size of the dataset increases. But here each entropy estimate is obtained from a dataset size of 182 (14x13) measurements at each time instant. That is very small, and none of the synthetic data tests made use of such small size in the verification of the results.

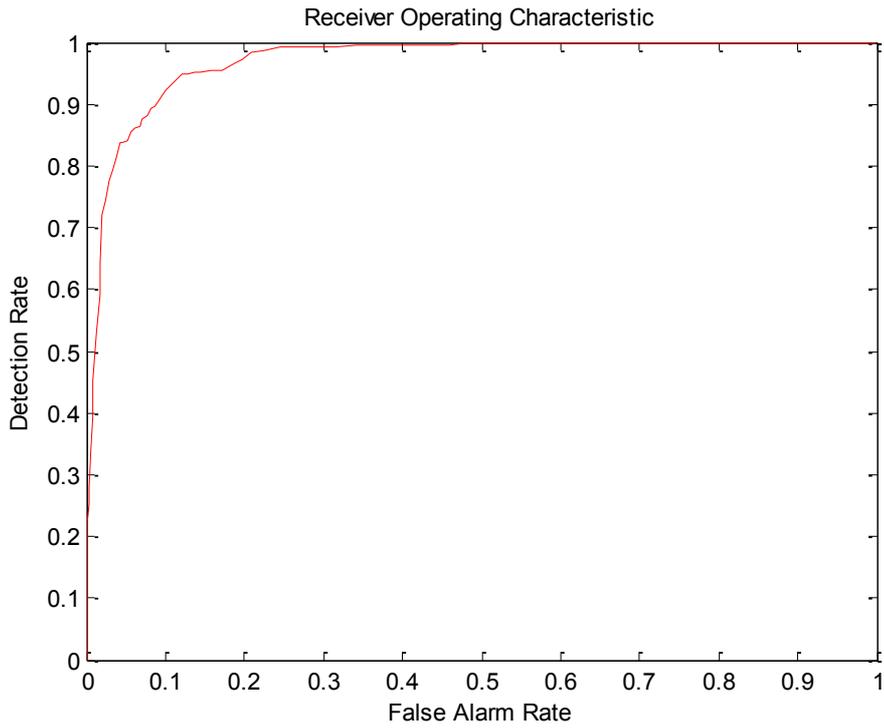

*Figure 3.36* *ROC plot for one-dimensional real data*

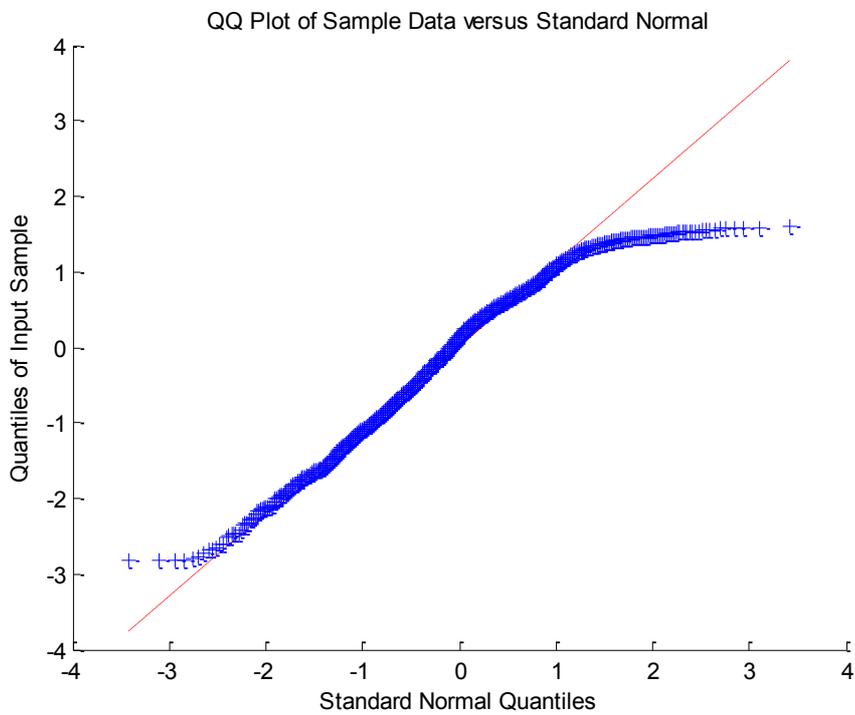

*Figure 3.37* *q-q plot of the Shannon entropy estimate.*

This result would be much better, and exactly as suggested in [1] if the dataset were large enough. The implication of this for monitoring wireless sensor network data is that the number of sensors needed to effectively detect anomalies will have to be very large, maybe approaching infinity to be very certain. This is another downside of this method. That said, many of the points still lie on the straight line anyway in this particular scenario. What is certain to happen in a situation where the number of sensors is limited in some way?

A good question that one might ask from observing the real data is; How does this system detect spatial anomalies? For example, one of the sensors is broken. The experiment was definitely not set up for that. The pdf of data obtained from sensor 1 and sensor 40 were estimated and the result is shown in Figure 3.38.

The q-q plot for spatial evaluation of the entropy of the data gave the result in Figure 3.39. By the sheer increase in the number of data points, the q-q plot provides a better result.

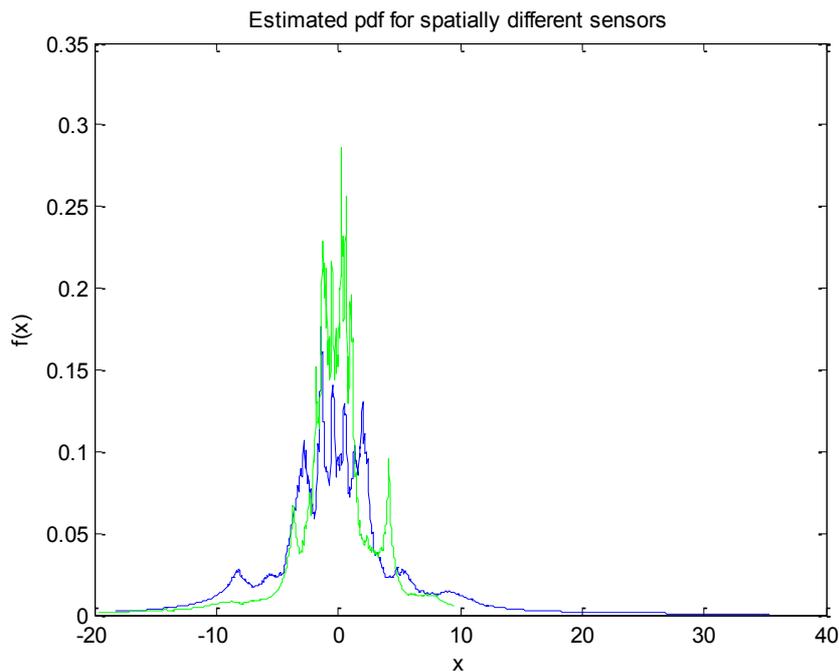

*Figure 3.38 Estimated pdf of the real data for sensors at different points*

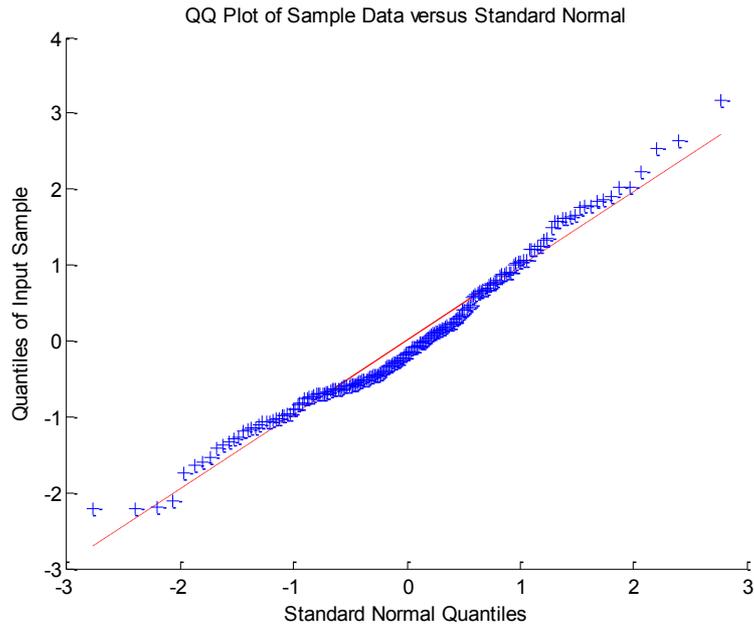

***Figure 3.39*** *q-q plot of the Shannon entropy estimate.*

### 3.8 Conclusions

From the foregoing discussion, the following are deduced:

The method of [1] is verified as shown by the results that have also been obtained and presented in this report. In fact, the method works well for all the distributions presented except for the Gaussian distribution in two dimensions, this of course, is due to its pdf being unbounded. The effects are not obvious in the one-dimensional Gaussian, and thus this method may still be safely applied with some caution. The entropy estimates get much better as the dataset sizes increase; this is shown by the reduction in the variance of the estimates with this increase. Also, the anomalies that can be detected by this method need to be very large compared with the actual measurement. This may not be the case in another real life scenario. Finally, the experiment seemed to have not been set up to also detect spatial anomalies. How can such anomalies be detected? A novel method is presented in the next chapter.

# CHAPTER 4
## 4.0 NOVEL APPROACH TO ANOMALY DETECTION
### 4.1 Introduction

The impact of subtle anomalies is completely missed by the method in [1]. The anomalies it seems to be able to detect are those with large RSSI measurements where one or more values are zero or very close to zero, or the other way round where we have sudden spikes in a measurement set. What happens in the case of small anomalies or subtle sabotage e.g. node replication, or inflation of data cannot be accounted for. This is because the entropy estimate which is used involves averaging the effect of the anomaly over the whole set. A possible solution will be any method which allows one to consider either some part or some section of the pdf alone or even particular points in the pdf. This line of thought was alluded to by [29] which uses the Kullback-Leibler divergence metric. This will be investigated in the following sections.

An example of the problem is shown in Figure 4.1 below. The datasets are Gaussian with exact same parameters except that one is zero mean and the other has mean of two. One is regarded as normal, while the other is regarded as anomalous.

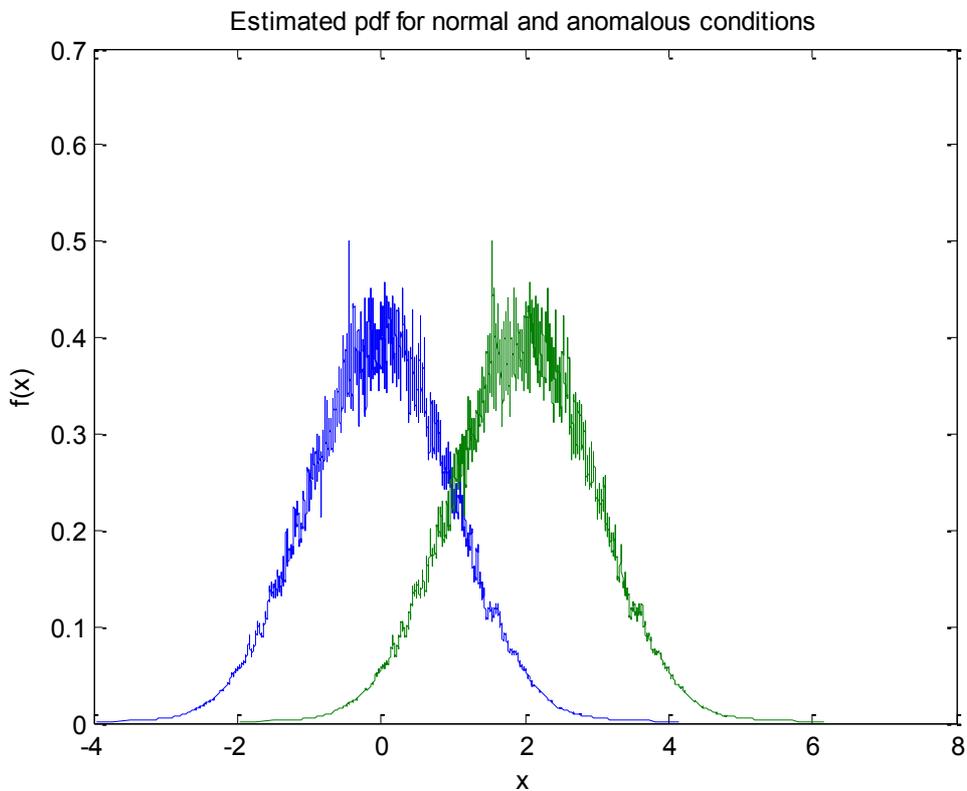

*Figure 4.1 Estimated pdf of two Gaussians with shifted means*

An evaluation of the entropy of the data for datasets of size 10000 gives 1.4203 and 1.4203 for both datasets respectively, exact same result. How do we threshold the method of [1] to spot an anomaly of this form? The theoretical value calculated from equation (11) gives 1.4189, really close to the estimated values.

It appears that apparently the method cannot handle such situations and will always miss a true detection. This problem is what this method seeks to solve. This chapter presents a novel method using the Bhattacharyya distance that would be able to detect such subtle anomalies, doing much better than [1]. It would be tested for Gaussian and beta distribution data in one dimension which would also be shown to be undetectable by [1], as well as for real data which [1] already said is normal.

### 4.2 Method

The suggested method brings together ideas from spectral analysis [30] as well as Bhattacharyya distance measure which has until now been used for classification. [31] [32] [33]

#### 4.2.1 Bhattacharyya Distance

This distance measure between discrete probability distributions was developed as an improvement on the Mahalanobis distance which works well for different covariance but becomes zero when they are equal. This becomes more obvious if two distributions are considered with only a shift in the mean of one from the other. This is not necessarily a bad thing for the Mahalanobis distance except that it is less robust that the Bhattacharyya distance. Thus intuitively, the latter would perform better with subtle anomalies. The Bhattacharyya Distance is defined as follows;

$$B_{ij} = \frac{1}{8}(m_i - m_j)^T \left(\left|\frac{(\Sigma_i + \Sigma_j)}{2}\right|\right)^{-1}(m_i - m_j) + \frac{1}{2}\ln\left(\frac{\left|\frac{(\Sigma_i + \Sigma_j)}{2}\right|}{\sqrt{(|\Sigma_i||\Sigma_j|)}}\right) \quad (15)$$

Where $m_i, m_j, \Sigma_i, \Sigma_j$ are mean and covariance of the dataset $i$ and $j$ respectively.

This is much easier to visualise in one dimension, where the mean is a scalar and the covariances are replaced by the variances of the datasets. It should also be noted that the dataset referred to here is that of the pdf that has already been estimated using [1]. It can also be observed from equation (15) that when the means are equal, the first term is zero, while when the variances are equal, the second term is zero.

For the example shown in Figure 4.1, theoretically, using the entire dataset at once, this evaluates to 0.5. This is a difference at least, unlike the method in [1] which returns exactly the same result for both datasets. This difference is amplified when the dataset is further divided into subsets so as to know where the largest differences exist.

### 4.3 Windowed Bhattacharyya Distance

Another new concept is introduced; a windowed Bhattacharyya distance. The main reason for this is that for a large dataset, anomalies are usually small compared to the actual dataset, so they may not be detected by measures that average the whole set. The windowed Bhattacharyya distance takes the form of placing a rectangular window on the pdf at the point of interest in both sets, and they are then compared for the discrete points covered in both sets. With this information, the method proceeds as follows.

### 4.4 Anomaly Detection using Bhattacharyya Distance

This method is described as follows: Given that the data for the normal condition is available, its pdf is evaluated with the $k$-nn method which has already been described in Chapter 3. When a new dataset is returned (test set), first, the pdf of the test set is also evaluated using the $k$-nn method. Then, the two pdf sets, the normal and the abnormal are compared based on the ideas from the last section.

To detect anomalies within the test set, a rectangular window divides the pdf estimate in both sets to subsets of equal lengths for which the Bhattacharyya distance is then calculated.

In order to determine where the window is placed, both datasets are compared, and the minimum and maximum of the two sets of the actual sensor measurements determine the range of the x-axis. The window starts at the minimum of both sets, and the other set is zero at this point and every other point for which it has no value whereas the other set has values.

The length of the window is arbitrarily chosen to be $k$.

### 4.5 Simulation using test data

The following are the results of the simulation carried out for the one dimensional Gaussian and beta distributions.

### 4.5.1 One dimensional Gaussian distribution

For the one dimensional Gaussian, data was generated in matlab. Two datasets, one normal and one, anomalous were tested. The normal condition is shown in blue in the figure below while the anomalous is shown in green. This kind of anomaly occurs when one inflates data, or reduces data by a uniform amount. One Gaussian is zero mean while the other has a mean of two. After estimating the pdf, the windowed Bhattacharyya distance was calculated for the entire range of data.

The result is as shown in Figure 4.2. This is consistent with the difference in the distributions that can be observed in the Figure 4.1.

Data in the range from -2 to -6 show higher distances than data from -2 to +2, the same trend seems to repeat itself on the positive side too. This is especially more obvious as the window size reduces. The effect of the reduction in window size can be seen in the Figure 4.3 for a window size of 20 and a dataset size of 10,000. This will be very much more sensitive to subtle anomalies and the threshold can be set as desired.

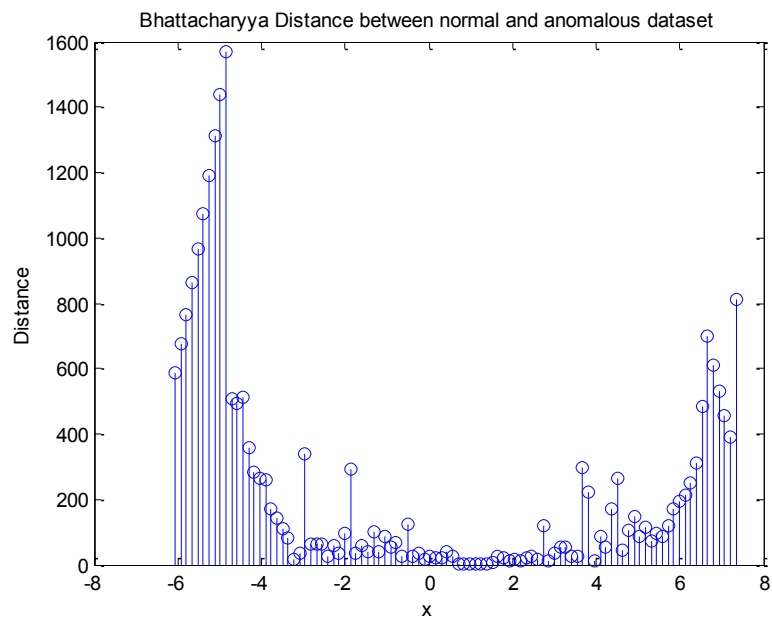

*Figure 4.2* Bhattacharyya Distance for the two Gaussians

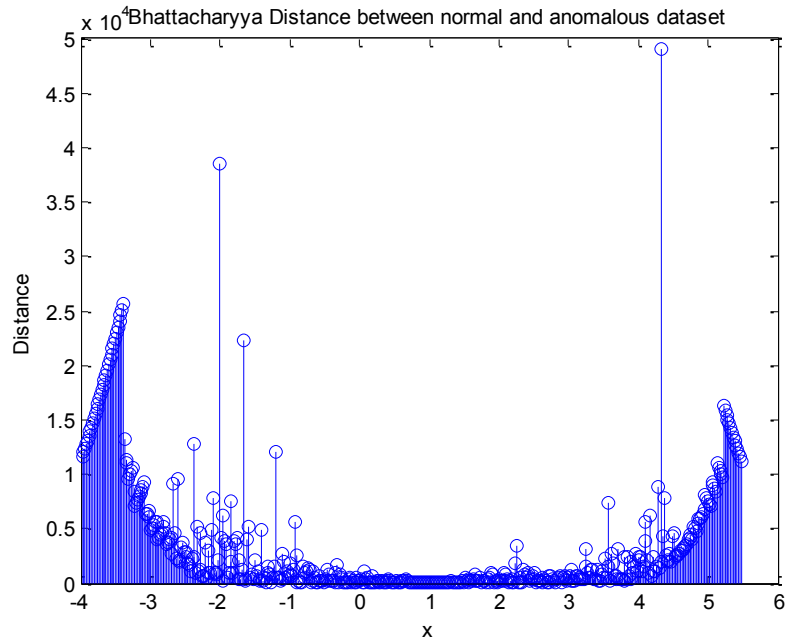

*Figure 4.3 Bhattacharyya Distance for two Gaussians with reduced in window size*

It is intuitive to expect this distance to increase as the shift in the means of the distributions increase.

### 4.5.2 One dimensional Beta distribution

The test is repeated for data from one dimensional Beta distribution. The pdf is shown below. By applying [1], the entropy of both sets are -0.3744 and -0.3744 respectively. By applying the windowed Bhattacharyya method, the result is shown in Figure 4.5.

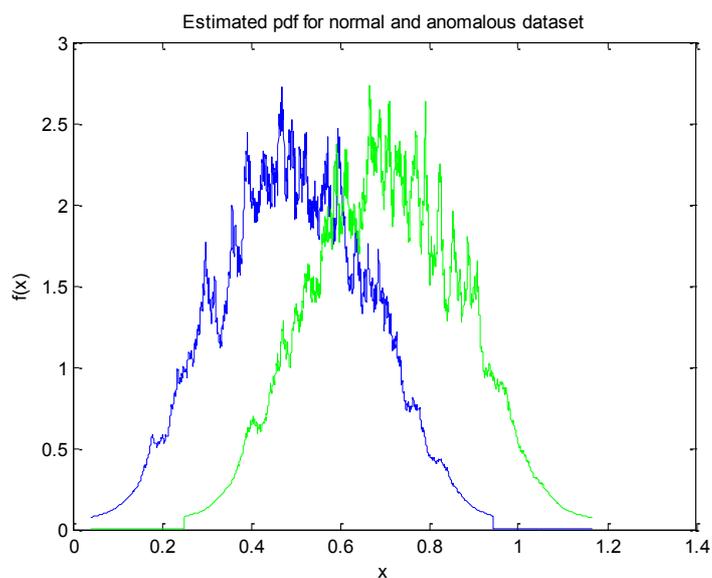

*Figure 4.4 Estimated pdf of two Beta Distribution with shifted means*

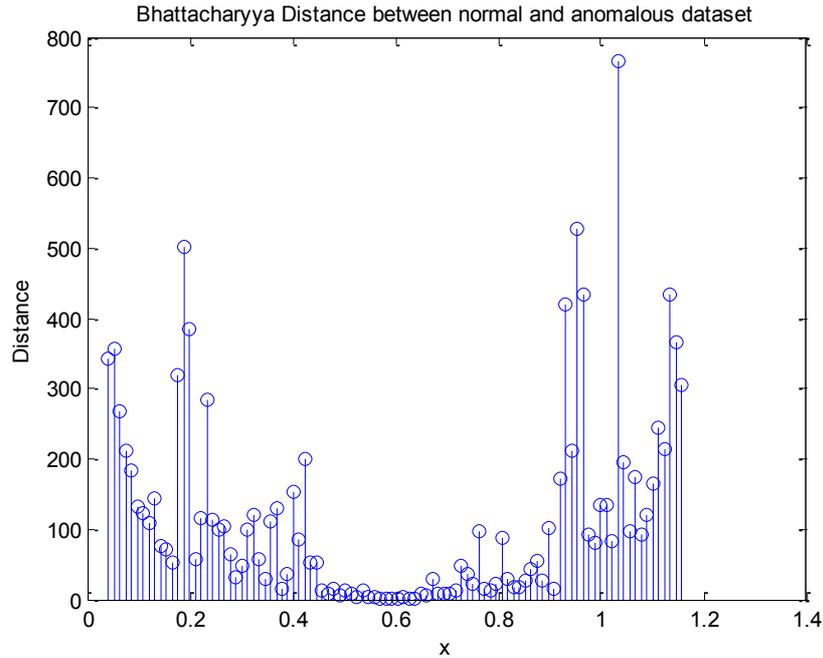

*Figure 4.5 Bhattacharyya Distance for the two Beta Distribution*

What interpretation can be given to these results? To better understand this, it may be useful to consider the following scenario(s). There are several sensors monitoring a particular variable(s). The normal conditions are known or have been estimated using the method of [1], then a dataset is obtained which maybe, takes the form of one of the distributions in Figure 4.1 and Figure 4.4, or any other form at all, for that matter. The two distributions are then divided into subsets of equal sizes and compared. Firstly, this method is better than the previous method because it may help analysts to know what range of data is reliable. That is, one may decide, based on the results of Figure 4.5 to neglect measurements higher than 0.9 if the normal is the one shown in blue, or maybe values within a range if the distance measure spikes within the dataset. This may be for both temporal and spatial measurements analysis.

### 4.6 Simulation using real data

The same method was applied for the real data, and as shown below, the threshold when moved appropriately would detect data that is just a little above the normal as seen in points around -28 and +20 in the Figure 4.6. As against the method of [1] where the anomalies have to be big enough to be spotted, here the method can detect slight increases, and this is expected to get better as the window size decreases. How much decrease can be accommodated by each application is however, another question entirely.

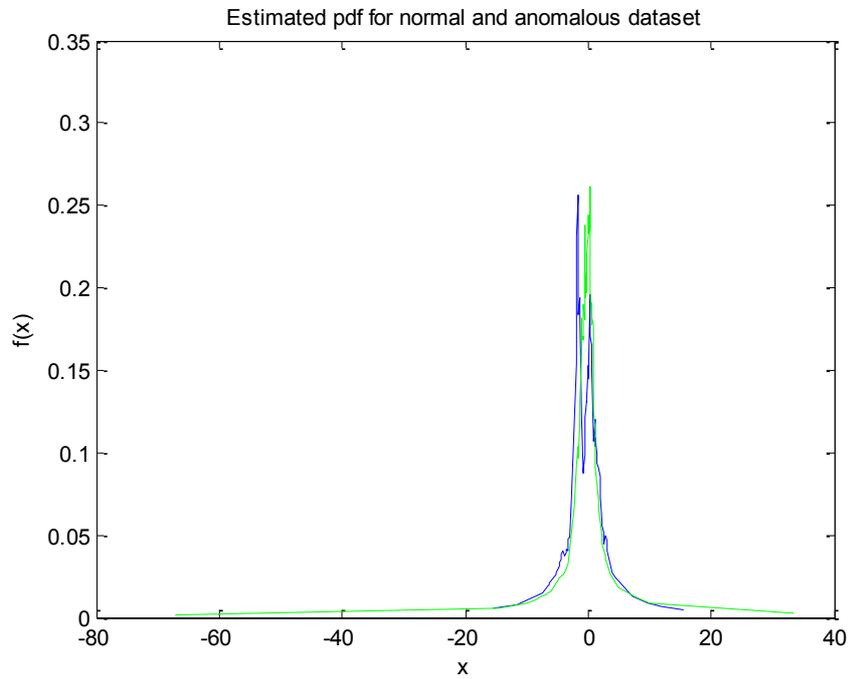

*Figure 4.6* *Estimated pdf of the real data at two different times.*

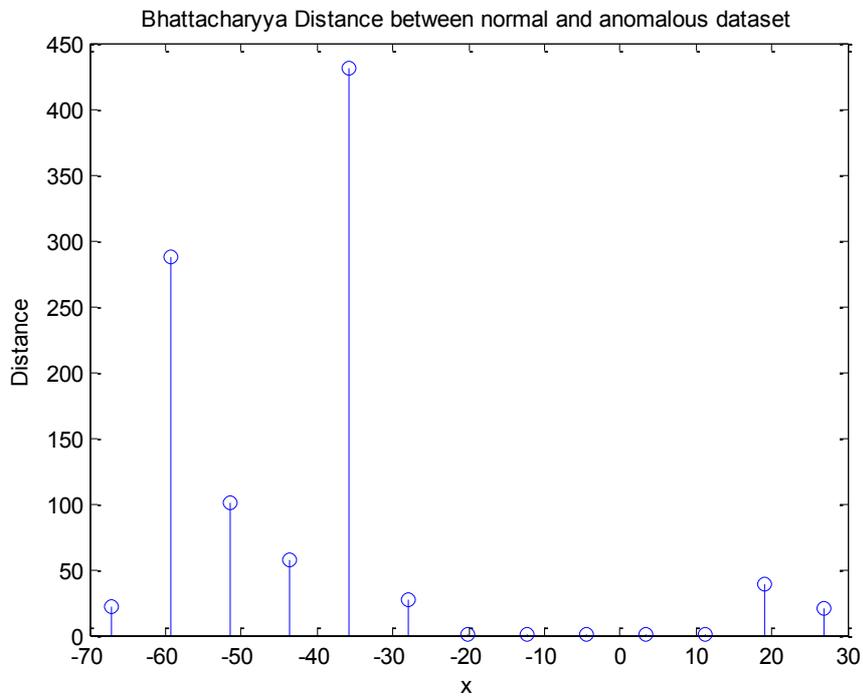

*Figure 4.7* *Bhattacharyya Distance for the real data at two different times*

## 4.7 Anomaly Detection using $k$-nearest neighbours plus Kullback-Leibler Divergence

A combination of the methods of [1] and [29] which has to the best of the writer's knowledge also not been done in the past gives another method of Anomaly detection. The pdfs are estimated using the $k-$nearest neighbour method while the Kullback-Leibler divergence is then used to compare the distributions. The difference between this method and [29] is that the authors of [29] used the histogram method for estimating the pdf. They had uniform bin size. Whilst [29] shows that the KL divergence is a very viable way to detect subtle anomalies, [1] shows that the k-nearest neighbour method is efficient for estimating the pdf. Both these can be consulted to further understand this.

# CHAPTER 5
## 5.0 CONCLUSIONS AND RECOMMENDATIONS
### 5.1 Conclusions

The method in [1] has been verified for all the distributions considered except the two dimensional Gaussian. The results presented are consistent with those in [1]. However, the shortcomings of [1] have been identified and a way to overcome this has been presented, tested for some test distributions, and also for the real data hosted at [28]. This method has been shown to perform better than the entropy based method in that it can detect more subtle anomalies. Although the results for the new method are presented for only one-dimensional data, the writer of this report believes that this method will outperform the entropy-based method in any dimension. Suggestions for future research are presented.

### 5.2 Recommendations for Future Research

The scalability of the novel method has not been investigated. It would be useful to verify that this works better than [1] also for higher dimensions even though that is what this writer believes.

Interpretability has not formed the crux of this discussion, it would be important to investigate this in wireless sensor networks as it will definitely make life easier for the analysts and decision makers.

# ACKNOWLEDGEMENTS

I am very grateful to God the giver, preserver and sustainer of life who has made this possible. I can do nothing without His help. To Him alone be all the glory and honour.

I would like to appreciate the Nigerian Government who sponsored my MSc studies through the Presidential Special Scholarship for Innovation and Development (PRESSID).

My sincere appreciation goes to my supervisor, Professor Bernard Mulgrew who still makes time out of his very busy schedule to be a very good supervisor. I would like to be like him when I grow up.

I thank my parents for their training and discipline that has helped me stay on course.

Thanks to everyone else who has contributed in one way or the other to make my studies in the University of Edinburgh a memorable one.

# APPENDIX

## Some Matlab Codes

Boundary-corrected pdf

```
al=4;be=4;
a = betarnd(al,be,10000,1);
f = 0.5;
N = floor(f*size(a,1)); %Split the Data
M = round((1-f)*size(a,1));
k = ceil(sqrt(N));
a1 = sort(a(1:M));
a2 = sort(a((M+1):end));
c1 = a1(:);
d1 = a2(:);
dx = min(diff(sort(c1)));
c1max = max(c1); c1min = min(c1);

Nn = round((c1max-c1min)/dx);
c2 = linspace(c1min,c1max,Nn);
%initialize the variables
d_distance = zeros(1,N);
fx = zeros(1,Nn);

for j = 1 : Nn
    dist = sort((c2(j) - c1).^2 ); %finds the distance to all other points
    d_distance(j) = sqrt(dist(k));
    if (d_distance(j) <= c2(j))
        vol = pi.* (d_distance(j)); %value of a sphere volume in one dimension
        fx(j) = (k-1)./(N*vol);
    else
    end
end
xfx = [c2 fx];
ind1 =  find((xfx(:,2) == 0));
ind2 =  find((xfx(:,2) ~= 0));
RS = length(ind2);
c1_nonzero = zeros(RS,1);
for rs = 1:RS
    ds = ind2(rs);
    c1_nonzero(rs) = d1(ds);
end

RR = length(ind1);
for rr = 1:RR
    dd = ind1(rr);
    dist = sqrt(((d1(dd) - c1_nonzero).^2));
    [MIN, IND] = min(dist);
    ind3 =  find((c1 == c1_nonzero(IND)));
    xfx(dd,2) = xfx(ind3,2);
end

ffx = sum(fx.*dx);
fx = fx./ffx;
figure, plot(c2,fx);
```

Plots
```
R = 50; %R = number of realisations
y = 90;
B = 10000;
Hh = zeros(y,1);
H_est = zeros(R,1);
H_st = zeros(y,1);
H_theor = zeros(R,1);
al =4; be =4;
H_Var = zeros(1,y);
sig = 2;
 for b = 1:y
     B = b*400;
     for i = 1:R
        a = sqrt(2)*randn(B,2); f = 0.5;
        N = f*size(a,1); %Split the Data
        M = (1-f)*size(a,1);
        k = ceil(sqrt(N));
        a1 = [(a((1:f*size(a,1)),1))  (a((1:f*size(a,1)),2))];
        a2 = [(a(((f*size(a,1)+1):end),1))  (a(((f*size(a,1)+1):end),1))];
        c1 = a1(:,1);
        d1 = a2(:,1);
        c2 = a1(:,2);
        d2 = a2(:,2);
%        initialize the variables
        c_distance = zeros(N,1);
        fx = zeros(N,1);
        for j = 1 : N
            dist = sort(((d1(j) - c1).^2 ) +((d2(j) - c1).^2));
%finds the distance to all other points
            c_distance(j) = (dist(k));
             if ((c_distance(j) <= d1(j)) && (c_distance(j) <=d2(j)))
                vol = pi.* (c_distance(j)); %value of a sphere volume
in one dimension
                fx(j) = (k-1)./(N*vol);
             else
            end
        end
        xfx = [d1 d2 fx];
         ind1 =  find((xfx(:,3) == 0));
         ind2 =  find((xfx(:,3) ~= 0));
         RS = length(ind2);
         d1_nonzero = zeros(RS,1);
         d2_nonzero = zeros(RS,1);
         for rs = 1:RS
             ds = ind2(rs);
             d1_nonzero(rs) = d1(ds);
             d2_nonzero(rs) = d2(ds);
         end
         
         RR = length(ind1);
         for rr = 1:RR
             dd = ind1(rr);
             dist = sqrt(((d1(dd) - d1_nonzero).^2) + ((d2(dd) - d2_nonzero).^2));
             [MIN, IND] = min(dist);
             ind3 =  find((d1 == d1_nonzero(IND)));
             xfx(dd,3) = xfx(ind3,3);
         end
%  fx = fx./ffx;
```

```
        G = (-log(fx)); %entropy functional
        H_est(i) = mean(G);
    end
    H_Var(b) = var(H_est);
    H_st(i) = mean(H_est);
end
x = 1:y;
H_stp = zeros(size(H_st));
H_stn = zeros(size(H_st));
H_ss = zeros(size(H_st));
for w = 1:y
    H_stp(w) = H_st(w) + 1.96*(sqrt(H_Var(w)./y));
    H_stn(w) = H_st(w) - 1.96*(sqrt(H_Var(w)./y));
     H_ss(w) = ((H_st(w) - mean(H_st))/sqrt(var(H_st)));
end

figure, plot(x, H_st, x,H_stp,x,H_stn);
title('Confidence intervals'),  xlabel('Sample Size'), % x-axis label
ylabel('H'), % y-axis label
```

ROC Plot

```
minH = min(H_ss);
maxH = max(H_ss);
ThNum = linspace(minH,maxH,R);
bb = length(1:30:R);
False = zeros(bb,1);
True = zeros(bb,1);
motionCode1 = motionCode(51:3178);
indd = find(motionCode1==0);
inddd = find(motionCode1==1);
for i = 1:30:R
    [ind] = find(H_ss>ThNum(end-i));
    False(i) = (length(intersect(indd,ind)))/(length(indd));
    True(i) = (length(intersect(inddd,ind)))/(length(inddd));
end
Record = [True False];
Record = sortrows(Record);
True = Record(:,1);
False = Record(:,2);
figure,  plot(False,True, 'r');
title('Receiver Operating Characteristic'),  xlabel('False Alarm Rate'), % x-axis label
ylabel('Detection Rate'), % y-axis label
```

Real Data Scan Statistics and q-q plot

```
ai = Z';
R = size(ai,2); %R = number of realisations
H_st = zeros(R,1);
    for i = 1:R
        a = ai(:,i); f = 0.5;
        N = floor(f*size(a,1)); %Split the Data
        M = round((1-f)*size(a,1));
        k = ceil(sqrt(M));
        a1 = sort(a(1:M));
        a2 = sort(a((M+1):end));
        c1 = a1(:);
        d1 = a2(:);
```

```matlab
        %initialize the variables
        d_distance = zeros(1,N);
        fx = zeros(N,1);
        for j = 1 : N
            dist = sort((d1(j) - c1).^2 ); %finds the distance to all other points
            d_distance(j) = sqrt(dist(k));
                vol = 2.* (d_distance(j)); %value of a sphere volume in one dimension
                fx(j) = (k-1)./(N*vol);
        end
        G = (-log(fx./ffx)); %entropy functional
        H_st(i) = mean(G);
    end
H_est = mean(H_st);
H_Varn = var(H_st(:));
H_stp = zeros(size(H_st));
H_stn = zeros(size(H_st));
H_ss = zeros(size(H_st));
for w = 1:R
    H_stp = H_est + 1.96*(sqrt(H_Varn./N));
    H_stn = H_est - 1.96*(sqrt(H_Varn./N));
    H_ss(w) = (H_st(w) - H_est)/(sqrt(H_Varn));
end
x = 1:R;
motionCode1 = motionCode(51:3177);
figure, qqplot(H_ss);
H_ss = H_ss./max(H_ss);
figure, plot(x, H_ss, x,motionCode1);
title('Scan Statistic'),  xlabel('Sample number'), % x-axis label
ylabel('H'), % y-axis label
```

Bhattacharyya Distance

```matlab
        al =4;be =4;
%        B = 10000;
        Z = Z';
        ggg = Z(:,1);
        [Xs, SortVec] = sort(ggg);
        UV(SortVec) = ([1; diff(Xs)] ~= 0);
        a = Z(1,:);
%          a = (Z(:,1));
        gga = (Z(:,40));
        [Xsa, SortVect] = sort(gga);
        UVa(SortVect) = ([1; diff(Xsa)] ~= 0);
        ai = Z(15,:);
        f = 0.5;
         B= length(a);
        N = floor(f*size(a,2)); %Split the Data
        M = round((1-f)*size(a,2));
        k = ceil(sqrt(N));
        a1 = sort(a(1:M));
        a2 = sort(a((M+1):end));
        c1 = a1(:);
        d1 = unique(a2(:));
        %initialize the variables
        d_distance = zeros(1,N);
        c_distance = zeros(1,N);
        doi = length(d1);
```

```matlab
            fx1 = zeros(doi,1);
            for j = 1 : doi
                dist = sort((d1(j) - c1).^2 ); %finds the distance to all other points
                d_distance(j) = sqrt(dist(k));
                vol = 2.* (d_distance(j)); %value of a sphere volume in one dimension
                fx1(j) = (k-1)./(N*vol);
            end
            fx1 = fx1/ffx;
            mind = min([min(d1) min(d2)]);
            maxd = max([max(d1) max(d2)]);
            d = linspace(mind,maxd,B);
            fx11 = zeros(B,1);
            fx22 = zeros(B,1);
            figure, plot(d1,fx1);
            for r = 1:B
                if (d(r)>=min(d1))&&(d(r)<=max(d1))
                    fx11(r) = interp1(d1,fx1,d(r),'spline');
                else
                    fx11(r) = 0.00;
                end
            end
            N = floor(f*size(ai,2)); %Split the Data
            M = round((1-f)*size(ai,2));
            a3 = sort(ai(1:M));
            a4 = sort(ai((M+1):end));
            c2 = a3(:);
            d2 = unique(a4(:));
            doo = length(d2);
            fx2 = zeros(doo,1);
            for j = 1 : doo
                dist = sort((d2(j) - c2).^2 ); %finds the distance to all other points
                d_distance(j) = sqrt(dist(k));
                vol = 2.* (d_distance(j)); %value of a sphere volume in one dimension
                fx2(j) = (k-1)./(N*vol);
            end
            fx2 = fx2/ffx;
            figure, plot(d2,fx2);
            for r = 1:B
                if (d(r)>=min(d2))&&(d(r)<=max(d2))
                    fx22(r) = interp1(d2,fx2,d(r),'spline');
                else
                    fx22(r) = 0.00;
                end
            end
%           k = 20;
 siz = round(B/k);
Db = zeros(siz,1);
for i = 1:k:(B-k)
    x1 = fx11(i:k+i-1);
    x2 = fx22(i:k+i-1);
    m1 = mean(x1);
    m2 =mean(x2);
    sig1 = var(x1);
    sig2 = var(x2);
if (sig1==0)
    sig1 = 0.00002;
else
```

```matlab
end
if (sig2==0)
    sig2 = 0.00002;
else
end
    Db(i) = (1/8).*((m1-m2).^2).*(2/(sig1+sig2)) + 
(1/2).*log(((sig1+sig2)/2)./(sqrt(sig1*sig2)));
end
Dd = zeros(siz,1);
bq = zeros(siz,1);
for s = 1:siz-1
    bq(s) = d(s*k-k+1);
    Dd(s) = Db(s*k-k+1);
end
figure, plot(d1,fx1);hold on;
plot(d2,fx2, 'g');hold off; title('Estimated pdf for normal and 
anomalous dataset'),
        xlabel('x'), ylabel('f(x)'), % y-axis label% x-axis label
figure, stem(bq, Dd); title('Bhattacharyya Distance between normal 
and anomalous dataset'),
        xlabel('x'), ylabel('Distance'), % y-axis label% x-axis label
```